\documentclass[lettersize,journal]{IEEEtran}
\usepackage{amsmath,amsfonts}
\usepackage{algorithmic}
\usepackage{array}
\usepackage[caption=false,font=normalsize,labelfont=sf,textfont=sf]{subfig}
\usepackage{textcomp}
\usepackage{stfloats}
\usepackage{url}
\usepackage{verbatim}
\usepackage{graphicx}
\usepackage{subfig}

\usepackage{multirow}
\usepackage{booktabs}
\usepackage{threeparttable}
\usepackage{cite}
\usepackage{color}

\def\BibTeX{{\rm B\kern-.05em{\sc i\kern-.025em b}\kern-.08em
    T\kern-.1667em\lower.7ex\hbox{E}\kern-.125emX}}
\usepackage{balance}
\begin{document}
\title{A Lightweight NMS-free Framework for Real-time Visual Fault Detection System of Freight Trains}
    \author{Guodong Sun, Yang Zhou, Huilin Pan, Bo Wu, Ye Hu, Yang Zhang
    \thanks{
		Manuscript received XXX; revised XXX; accepted XXX. Date of publication XXX; date of current version XXX. 
		This work was supported in part by the National Natural Science Foundation of China (Grant 51775177) and PhD early development program of Hubei University of Technology (Grant XJ2021003801).
		The Associate Editor coordinating the review process was XXX. (Corresponding author: Yang Zhang.)
		
		G. Sun, Y. Zhou, H. Pan, Y. Hu and Y. Zhang are with School of Mechanical Engineering, Hubei University of Technology, Wuhan 430068, China (e-mail: sgdeagle@163.com; z-g99@hbut.edu.cn; hlp@hbut.edu.cn; huywfly@163.com; yzhangcst@hbut.edu.cn).

		B. Wu is with the Shanghai Advanced Research Institute, Chinese Academy of Sciences, Shanghai 201210, China (wubo@sari.ac.cn)
				
		Color versions of one or more of thefigures in this article are available online at XXX. 
		
		Digital Object Identifier XXX

    }}

    \markboth{September~2020}  
    {How to Use the IEEEtran \LaTeX \ Templates}

\maketitle

\begin{abstract}
	Real-time vision-based system of fault detection (RVBS-FD) for freight trains is an essential part of ensuring railway transportation safety. Most existing vision-based methods still have high computational costs based on convolutional neural networks. The computational cost is mainly reflected in the backbone, neck, and post-processing, i.e., non-maximum suppression (NMS). In this paper, we propose a lightweight NMS-free framework to achieve real-time detection and high accuracy simultaneously. First, we use a lightweight backbone for feature extraction and design a fault detection pyramid to process features. This fault detection pyramid includes three novel individual modules using attention mechanism, bottleneck, and dilated convolution for feature enhancement and computation reduction. Instead of using NMS, we calculate different loss functions, including classification and location costs in the detection head, to further reduce computation. Experimental results show that our framework achieves over 83 frames per second speed with a smaller model size and higher accuracy than the state-of-the-art detectors. Meanwhile, the hardware resource requirements of our method are low during the training and testing process.
\end{abstract}



\section{INTRODUCTION}
Nowadays, the efficiency and quality of braking systems for freight trains have become essential in ensuring transportation safety. The braking systems will seriously fail if some critical components are lost or damaged, such as bogie block key, dust collector, fastening bolt on brake beam, etc.
However, the faults of these components have been detected manually for a long time, resulting in missing and false detection. Recently, the applications of vision-based detection methods to replace manual labor can ensure the speed and accuracy of detection and avoid the subjective influence of manual detection.

Real-time vision-based system of fault detection (RVBS-FD) is one of the most important infrastructures in the intelligent transportation system. The RVBS-FD has been regarded as a critical technology in the instrumentation and measurement (IM) field~\cite{Zhang_TIM}.
The deployment of RVBS-FD is generally outdoors, as shown in Fig.~\ref{fig1}, and the computing resources are also limited. Moreover, due to the small size of these critical parts in trains, the structures are complicated and susceptible to contamination. In this case, the typical RVBS-FD cannot simultaneously meet the actual requirements of real-time detection and high accuracy.

With the advancement of machine learning in recent years, some scholars have developed several RVBS-FD that use machine learning as a detection method. 
For example, Liu et al.~\cite{4} used cascade detectors and gradient coded co-occurrence matrix to detect the faults of the bearing cap. 
Sun et al.~\cite{OE} proposed a fast adaptive Markov Random Field (FAMRF) algorithm and an exact height function (EHF) to detect the faults in the air brake system, bogie block key, and fastening bolt. Furthermore, Sun et al. compared the performance with the cascade detector based on local binary patterns (LBP).
Lu et al.~\cite{3} presented an automatic visual inspection for multiple targets based on the time-scale normalization to detect small mechanical parts.
Traditional RVBS-FD for freight trains has several unavoidable shortcomings in real-world applications, such as a restricted number of fault types and algorithm portability.

As deep learning algorithms are increasingly used in RVBS-FD, the key parts of freight trains that can be detected extend to valves~\cite{2}, bearings~\cite{Fu}, shaft bolts~\cite{Sun} and some small components~\cite{7, Ye, XiaoL}.
Zhang et al.~\cite{Zhang_TIM} used a multi-region proposal network and a multilevel region of interest algorithm based on convolutional neural network (CNN) to rapidly detect five common faults at the bottom of the train.
Fu et al.~\cite{Fu} proposed a two-stage network that could learn to automatically focus on the target defect area to detect bearing oil spots.
Chen et al.~\cite{5} proposed a two-stage detection method to achieve hierarchical detection and apply fine-grained defect diagnosis of train components.
However, the above methods have a large amount of calculation and slow detection speed, which cannot meet real-time detection requirements for over 30 frames per second (FPS). Furthermore, the computational cost of the network is often in the generation and suppression of anchor boxes. Inspired by OneNet~\cite{OneNet}, we use a one-stage network for detection without region proposal network and NMS post-processing. Moreover, the computational cost of the one-stage detector is mainly reflected in the backbone and neck. As a result, we improve these two parts to achieve higher accuracy and real-time detection at a low computational cost.

In this paper, we propose a lightweight NMS-free framework of vision-based fault detection for braking systems in freight trains. We first use a lightweight backbone for feature extraction and then design a fault feature pyramid (FFP) to obtain information across different scales. Finally, to avoid using NMS in post-processing, we put different scales information into the detection head and calculate different loss functions, including classification and location loss. 
In FFP, we design three modules for feature extraction based on the characteristics of the critical components in the braking systems of freight trains. These three modules are fault enhance attention, dilated fault bottleneck and fault bottleneck modules. The first module fits complex correlations between channels to improve the quality of representations produced by the network. The other two modules reduce the model parameters and calculations.
Experiments on three datasets of critical components in braking systems show that the detection speed of our framework is over 83 FPS, with a lower computational cost in comparison with the state-of-the-art detectors. Moreover, our framework has the smallest model size, suitable for scenarios with limited detection resources.

In general, our main contributions are as follows.
\begin{enumerate}
	\item{We design a lightweight NMS-free framework in the RVBS-FDet for freight trains.}
	\item{We proposed the FFP with three individual modules to improve the accuracy and significantly reduce the computational cost.}
	\item{The experimental results verify that our method achieves the detection speed of over 83 FPS with smallest model size in comparison with the state-of-the-art methods.}
\end{enumerate}

The rest of this article is organized as follows. Section~\ref{RELATED WORKS} introduces the related works about fault detection of freight train images, backbone, NMS-based object detection and NMS-free object detection. Section~\ref{HARDWARE SYSTEM} introduces the hardware system of our framework. Section~\ref{DETECTION  ALGORITHM} presents our overall framework and three individual modules. To verify the effectiveness of our method, we conduct comprehensive experiments in Section~\ref{EXPERIMENTS}. Finally, the full text is summarized in the Section~\ref{6}.

\section{RELATED WORKS}
\label{RELATED WORKS}
As technology advances, deep learning-based fault detection becomes more popular among RVBS-FDs. Recent research articles on RVBS-FD of freight trains and neural network construction are reviewed in this section.

\subsection{Fault Detection of Freight Train Images}
Sun et al.~\cite{Sun} proposed an automatic multi-fault recognition in the trouble of moving freight train detection system, which can identify four typical side frame keys and shaft bolts failures. 
Sun et al.~\cite{Sun.J} used binocular vision to detect bolt-loosening in key parts of the train. 
Ling et al.~\cite{Ling} proposed an instance detection model based on hierarchical features, which can perform instance-level prediction of rough defect areas. 
Ye et al.~\cite{Ye} proposed a multi-feature fusion network for simultaneous detection of three typical mechanical component failures, which has good robustness to complex environments. 
Pahwa et al.~\cite{2} proposed a method to detect valves failure using image segmentation followed by neural network recognition.
Chen et al.~\cite{5} proposed an end-to-end detection framework for fusing the visual features and substructure knowledge to detect the failure of train components.
Chang et al.~\cite{7} proposed fault detection and localization method for key components of high-speed trains.
However, the above methods have high computational costs and cannot meet the actual needs such as real-time detection. These methods are not suitable for deployment in a resource-limited environment in the wild.
\subsection{Backbone}
The backbone is usually used for the preliminary features extraction, generating feature maps for the subsequent network.
MobileNet~\cite{MoilbNetv3Small} used deep separable convolution, inverted residuals, linear bottleneck, and squeeze-and-excitation block in the network to achieve faster speed and higher accuracy.
The advantage of MobileNet was that the minimal model size could obtain faster speed, and some networks with the same characteristics had PeleeNet~\cite{Pelee}, MnasNet~\cite{MnasNet}, GhostNet~\cite{GhostNet}, and RegNet~\cite{RegNet}.
SqueezeNet~\cite{SqueezeNet} achieved the lightweight of the network by replacing the convolution and fire-module structure, reducing the number of parameters by 50 times.
Tan et al.~\cite{EfficientNet-B0} systematically studied model scaling and found that carefully balancing the network depth, width, and resolution of the network can improve performance.
He et al.~\cite{ResNet} proposed the residual net (ResNet), which used residual learning to solve the degradation problem to train a deeper network.
The model size will inevitably increase as the network deepens, but the accuracy will also increase, such as CSPDarkNet~\cite{CSPDarkNet-53}, deep layer aggregation (DLA)~\cite{DLA}, and VoVNet~\cite{VoVNet27-slim}.

\begin{figure*}[t]
	\begin{center}
		\centering
		\includegraphics[width=6.5in]{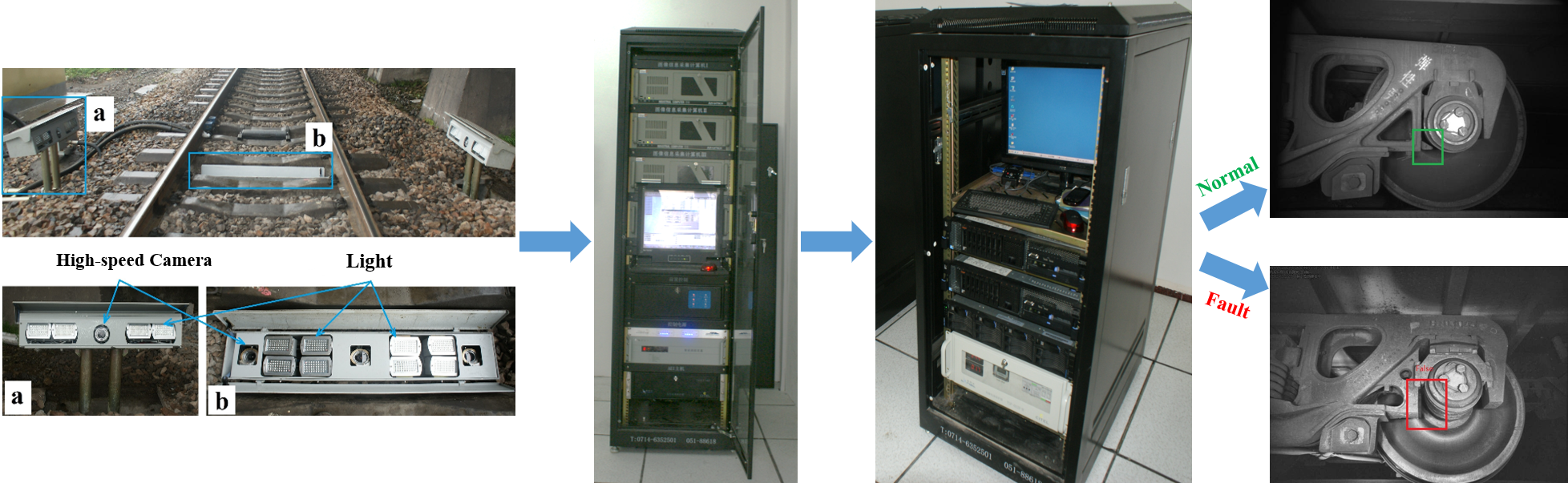}
		\caption{Hardware system for real-time vision-based fault detection framework of braking systems in freight trains. The system consists of the image acquisition subsystem, the data storage and analysis subsystem, and the image processing subsystem. The image acquisition subsystem includes a group of high-speed cameras and several lights, combined and placed on the bottom and both sides of the trains.}
		\label{fig1}
	\end{center}
\end{figure*}

\subsection{NMS-based Object Detection}
\subsubsection{Two-stage}
The proposal of Faster R-CNN~\cite{Faster_R-CNN} established the position of the two-stage methods in the field. Faster R-CNN was an accurate and effective region proposal network that shared convolutional features with downstream detection networks to improve overall accuracy. Lu et al.~\cite{Grid_R-CNN} proposed Grid R-CNN to achieve high-quality positioning through a grid guidance mechanism. 
Ze et al.~\cite{RepPoints} proposed the RepPoints to fine-grain localization information and identified local areas that were meaningful for object classification. Zhang et al.~\cite{Dynamic_R-CNN} proposed that the fixed scheme limits the overall performance and uses simple but effective components to bring improvements without additional costs. Cai et al.~\cite{Cascade_R-CNN} proposed Cascade R-CNN utilizing the output of one to train the next, and the same cascade was applied at inference.
The two-stage network generates potential object bounding boxes, called anchors. These boxes are adjusted and classified in post-processing, namely NMS. The generated anchor boxes are mapped to the areas of feature map, and then the areas are re-input to the fully connected layer for classification and regression. Each area mapped by the anchor must be classified and regressed, consuming significant times.
\subsubsection{One-stage}
The one-stage methods can directly regress and classify the anchor boxes. These methods have fewer parameters and calculations than the two-stage methods, which meet the fault detection requirements with low computational cost.
It was not until the emergence of single shot multibox detector (SSD)~\cite{SSD} and you only look once (YOLO) \cite{YOLOv3} that the one-stage methods began to attract attention.
The detectors of the same series of YOLO, such as YOLOF~\cite{YOLOF} and YOLOX~\cite{YOLOX}, had different improvements in network structure, which further promoted the development of one-stage detection methods.
CornerNet~\cite{CornerNet} used a single CNN to detect the target bounding box composed of two key points in the upper left corner and the lower right corner.
Fully convolutional one-stage object detection (FCOS)~\cite{FCOS} eliminated the pre-defined anchor box to avoid the calculation and related hyper-parameters during the training process. 

CentripetalNet~\cite{CentripetalNet} introduced a simple but effective centripetal shift to solve the corner matching problem. 
Xiang et al.~\cite{Generalized_focal_loss} proposed the generalized focus loss to encourage the learning of better classification and localization quality joint representations with more informative and accurate bounding box estimation.
Lin et al.~\cite{Focal_Loss} proposed the focal loss, which applied a modulating term for the cross entropy loss to focus learning on complex negative examples.
Zhang et al.~\cite{VarifocalNet} proposed VarifocalNet to learn an IoU-aware classification score as a joint representation of object presence confidence and localization accuracy, then designed a new loss function to train a dense object detector to predict the IoU-aware classification score.
\subsection{NMS-Free Object Detection}
Using NMS to eliminate redundant boxes increases calculations while reducing the detection speed. So, some methods avoid using NMS as the post-processing. For instance, Zhou et al.~\cite{Object} proposed CenterNet to achieve the bounding box based on the center point. All outputs were directly produced from the keypoints and did not need IoU-based NMS or any other post-processing.
Sun et al.~\cite{Sparse_R-CNN} proposed Sparse R-CNN for object detection in images. A fixed sparse set of learned object proposals were provided to perform classification and location by dynamic heads.
Zhu et al.~\cite{Deformable_DETR} proposed a multi-scale deformation attention module to improve the convergence speed. 
Pang et al.~\cite{Balanced} used IoU-balanced to reduce redundant frames and achieved similar effects as NMS to solve the imbalance through an overall balanced design.
Sun et al.~\cite{OneNet} introduced the concept of additional classification cost and scored gap to make the detector generate one-to-one predictions for removing the NMS to establish an end-to-end system.

\begin{figure*}[t]
	\begin{center}
		\centering
		\includegraphics[width=6in]{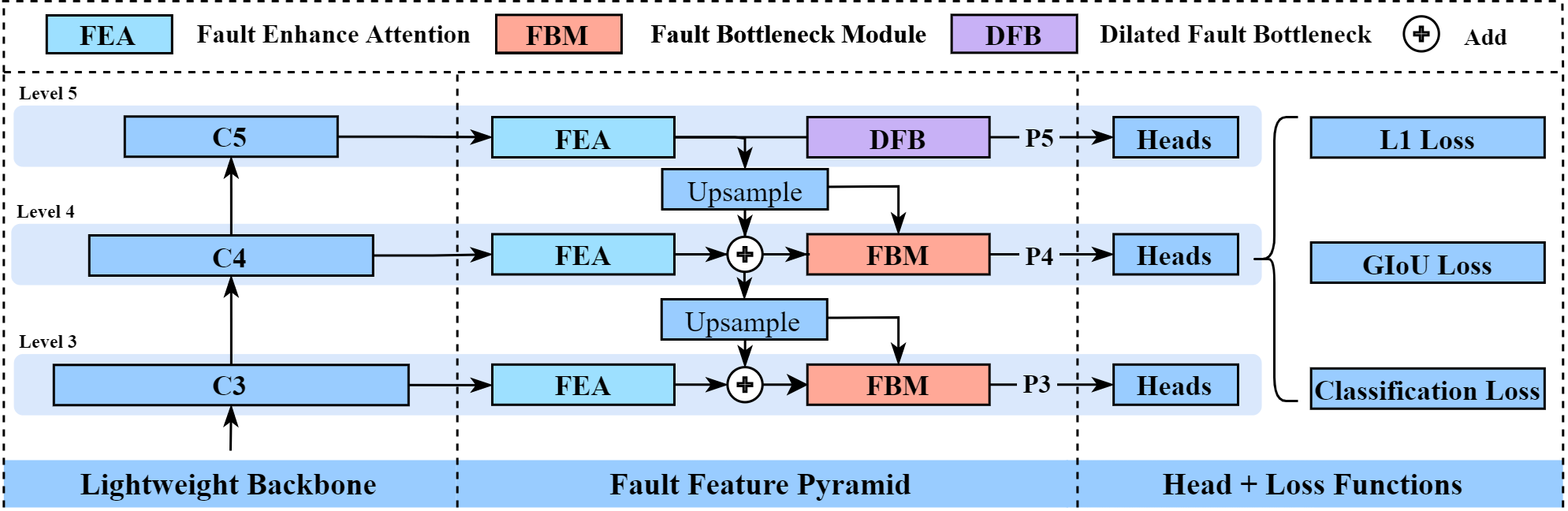}
		\caption{Overall network diagram of our proposed NR FTI-FDet. This network is divided into the backbone, fault feature pyramid (FFP), and detection head. The FFP contains fault enhance attention (FEA), dilated fault bottleneck (DFB) and fault bottleneck module (FBM). We calculate three different losses in the head to avoid using NMS in post-processing. The DFB only exists in the top layer, and the dilation rates are [1, 2, 5].}
		\label{fig2}
	\end{center}
\end{figure*}

\section{HARDWARE SYSTEM}
\label{HARDWARE SYSTEM}
As shown in Fig.~\ref{fig1}, the proposed RVBS-FD mainly includes an image acquisition subsystem, an image processing subsystem, and a data storage and analysis subsystem. The image acquisition subsystem is composed of a high-speed camera array and a light array~\cite{Zhang_TIM}. The image acquisition subsystem is spread at the bottom and on both sides of the track, eliminating the case that the image cannot be acquired due to the occlusion of parts. The high-speed camera array comprises several high-speed digital charge-coupled device cameras, and the size of the image collected each time is fixed at $700\times 512$ pixels. The light array comprises several xenon lamps, irradiated at different angles to avoid poor image quality due to uneven illumination. The image processing subsystem and the data storage and analysis subsystem are generally used to receive, process, and store the image to be tested returned by the image acquisition subsystem.
\section{DETECTION  ALGORITHM}
\label{DETECTION  ALGORITHM}
We propose a lightweight NMS-free framework for real-time vision-based fault detection of braking systems in freight trains (NR FTI-FDet). 
The proposed framework mainly includes a lightweight backbone, FFP structure, and detection head. The FFP structure consists of three modules: fault enhance attention, dilated fault bottleneck, and fault bottleneck module, as shown in Fig.~\ref{fig2}. 
In this section, we first introduce the overall framework and then the FFP with three modules. Finally, we present the detection head and loss function used in the network.

\subsection{Overall Framework}
\label{4.2}
The outdoor environment limits fault detection of freight train images. So that, the fault detector needs to be low computational costs. However, a large amount of calculation often occurs in the backbone, neck, and post-processing just introduced in OneNet~\cite{OneNet}. To reduce the computational costs, we use the lightweight backbone MobileNetV3-Small~\cite{MoilbNetv3Small}. Because some faults occur on small parts, the pixel information of the fault is gradually lost as the network deepens during the backbone processing. To better detect this type of fault with a large difference in size, we add a neck similar to FPN~\cite{FPN} in the back-end of the backbone. 

As shown in Fig.~\ref{fig2}, the backbone can generate the feature maps across different scales. High-level feature maps have richer semantic information, and low-level feature maps have higher resolution. We extract some critical feature maps and input them into the FPN. Through the processing of FPN, the rich semantic information of the upper-layer feature maps is transferred to the lower-layer high-resolution feature maps for fusion. Then the fusion feature maps are input into the detection head to detect faults in different sizes.

The addition of FPN can improve the detection accuracy and also leads to a substantial increase in computational costs, as shown in Table~\ref{Overall network structure} and Table~\ref{Fault Feature Pyramid}, respectively. In practical applications, we have to use a low computation and small model size detector and compromise accuracy~\cite{Zhang_TII},  limited by the resource in the wild. So we propose three individual modules to optimize the network for achieving a good trade-off between the accuracy and the amount of calculation.

Using NMS in post-processing to eliminate redundant boxes is the current practice of most two-stage and one-stage detectors. In searching for local maxima and suppressing non-maximum elements, NMS needs to calculate the IoU of two overlapping boxes, which significantly impacts detection speed. To solve the above problems, we calculate three loss functions to achieve one-to-one prediction, which will be discussed in detail in Section~\ref{Loss Function}.
\subsection{Fault Enhance Attention}
\begin{figure}[t]  
	\begin{center}    
		\subfloat[]
		{
			\includegraphics[height=1.6in]{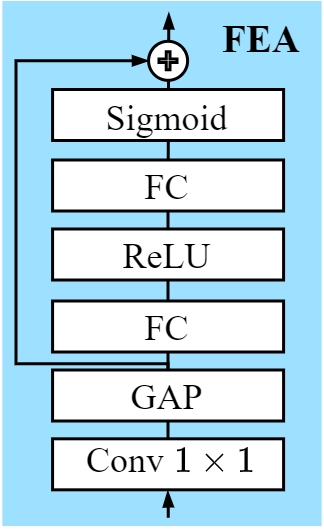}
			\label{1}
		}
		\subfloat[]
		{
			\includegraphics[height=1.6in]{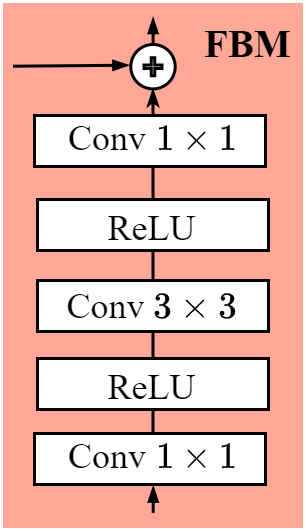}
			\label{2}
		}
		\subfloat[]
		{
			\includegraphics[height=1.6in]{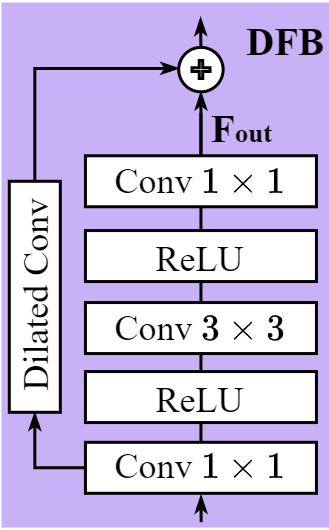}
			\label{3}
		}
		\caption{The proposed modules in fault feature pyramid (FFP). (a) Fault enhance attention (FEA). (b) Fault bottleneck module (FBM). (c) Dilated fault bottleneck (DFB). FC means fully connected layer, and GAP means global average pooling layer.}
		\label{modules}
	\end{center}
\end{figure}

In the traditional FPN~\cite{FPN}, the feature maps pass from the backbone are only processed by a $1\times 1$ convolution to adjust the number of channels. However, if the network does not model interdependencies among channels, the helpful information will be submerged in redundant information.
Therefore, we design a fault enhance attention (FEA) module to solve the above problem. As shown in Fig.~\ref{modules}(a), when the output feature map from the backbone enters into FEA, we use a $1\times 1$ convolution to adjust the channels to 256. We then obtain a $1\times 1\times 256$ feature map by the global average pooling. After that, two fully connected layers realize the non-linear operation among channels through upscaling and reducing the dimensionality of the feature map to fit the complex correlation among channels. Finally, the processed feature map is added with the original feature map to output after a sigmoid layer. We model the interdependence among channels and produce important values for subsequent processing through FEA.

\begin{figure}[t]  
	\begin{center}    
		\subfloat[]{\includegraphics[width=1.4in]{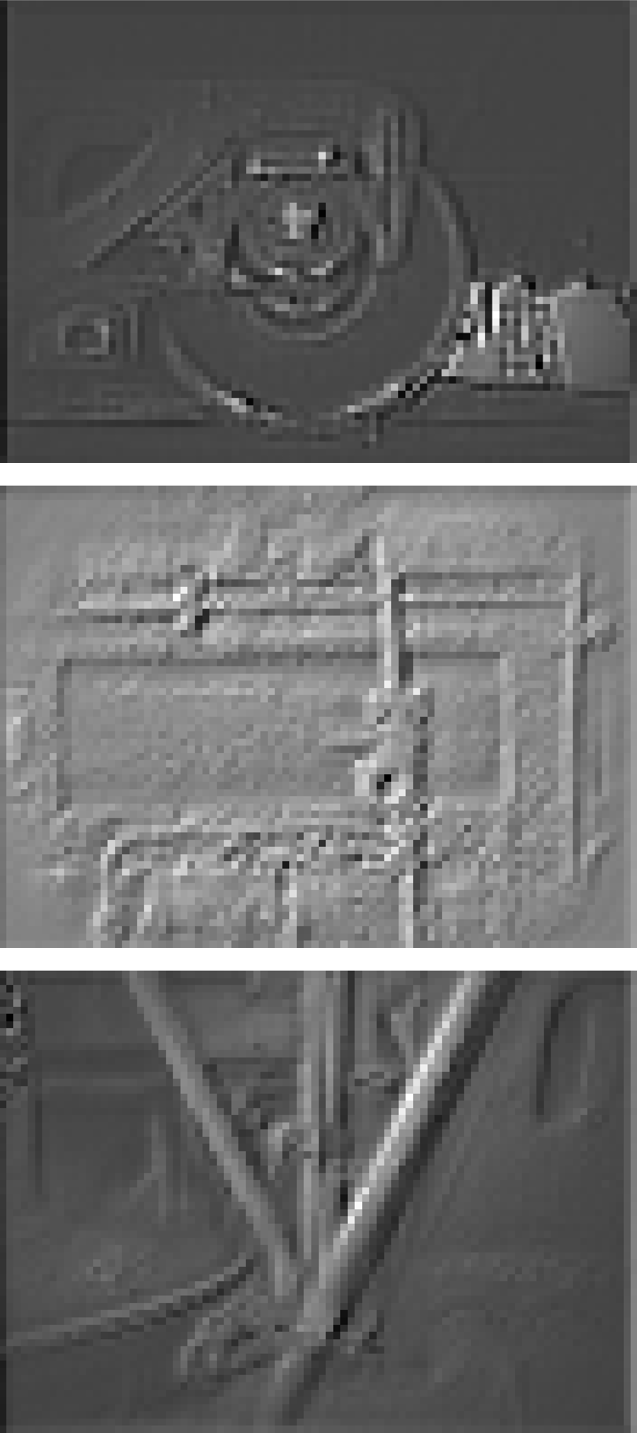}}
		\hspace{0.3em}
		\subfloat[]{\includegraphics[width=1.4in]{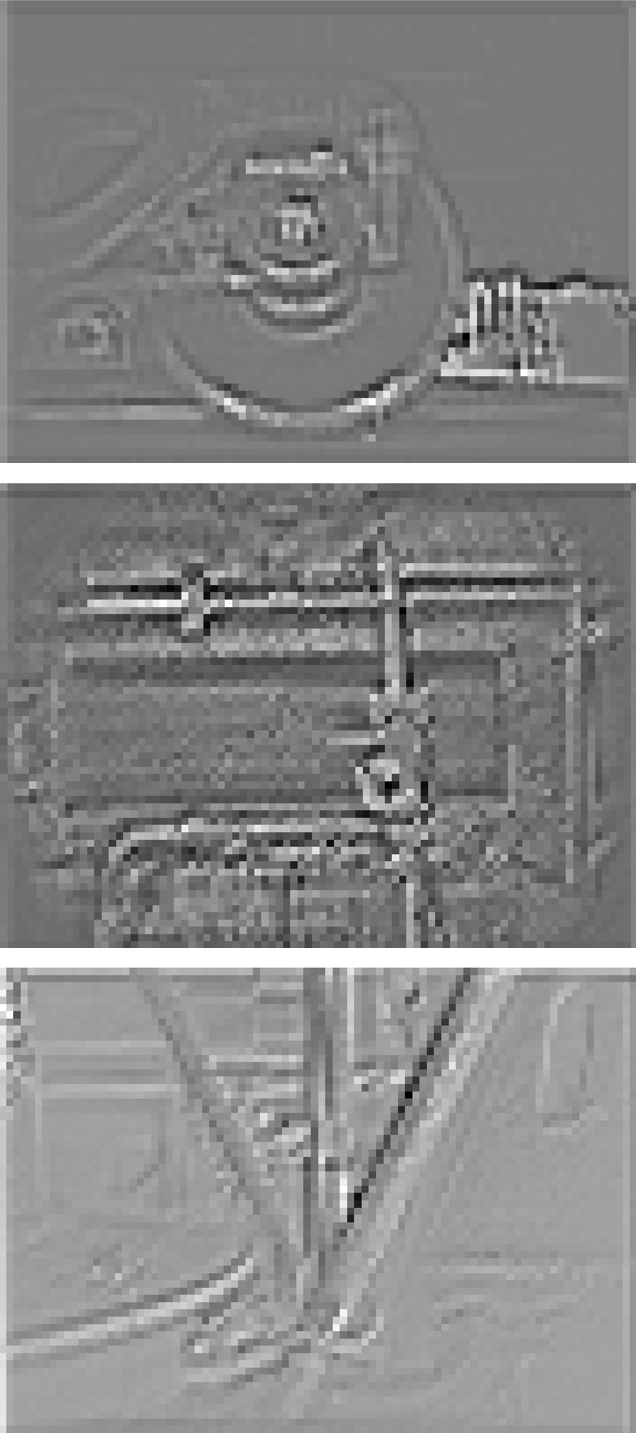}}
		\caption{Comparison of the average feature maps. (a) The feature map from C3 after a $1\times 1$ convolution. (b) The feature maps from C3 after the FEA module. The feature maps in (b) retain more textures and provide better fine-grained information for subsequent processing.}
		\label{P3}
	\end{center}
\end{figure}

We define an input as $x$, the FEA can be expressed as: 
\begin{equation}
	\label{FEA}
	{\rm{FEA}} = f\left( {g\left( x \right),x} \right) = g\left( x \right) \cdot x.
\end{equation}
The $g(x)$ is the transformation function of the input $x$, and $g(x)$ can be expressed as: 
\begin{equation}
	\label{FEA2}
	g(x) = Sigmoid\left( {MLP\left( {GAP\left( x \right)} \right)} \right).
\end{equation}
So the output $y$ is expressed as: 
\begin{equation}
	\label{FEA3}
	y = \sigma \left\{ {{W_1}\delta \left[ {{W_0} \cdot GAP\left( x \right)} \right]} \right\} \cdot x,
\end{equation}
where $\sigma$ is the sigmoid function, $W_0$ and $W_1$ are the multilayer perceptron weights, respectively. We compare the average feature maps after a $1\times1$ convolution with after FEA in Fig.~\ref{P3}. The average feature map processed by FEA retains more textures and provides better fine-grained information for subsequent processing.
\subsection{Dilated Bottleneck Module}
To further reduce the computational costs, we use $1\times 1$, $3\times 3$ and $1\times 1$ convolutions instead of the $3\times 3$ convolutions in the traditional FPN. The first $1\times 1$ convolution adjusts the channel from 256 to 16, and the last $1\times 1$ convolution adjusts the channel from 16 to 256. After that, we can significantly reduce the number of model parameters. If the number of channels in the input convolution layer is $C_i$, the number of output channels is $C_o$, and the convolution kernel size is $K\times K$, the number of parameters $P$ can be calculated as $P=C_i\times K^2\times C_o$. The parameter $P$ of our method is 10496, which is only $1/56$ of the traditional method.

To better use the semantic information from the upper layers, we add a shortcut to connect the upper layers and implement a residual-like structure~\cite{ResNet}, which is named fault bottleneck module (FBM) as shown in Fig.~\ref{modules}(b). The addition of the shortcut reduces the probability of gradient dissipation or explosion. Furthermore, the rich semantic information from the upper layer can be directly involved in the subsequent detection process of the network to improve the accuracy.
\begin{figure}[t]  
	\begin{center}    
		\subfloat[]{\includegraphics[width=1.4in]{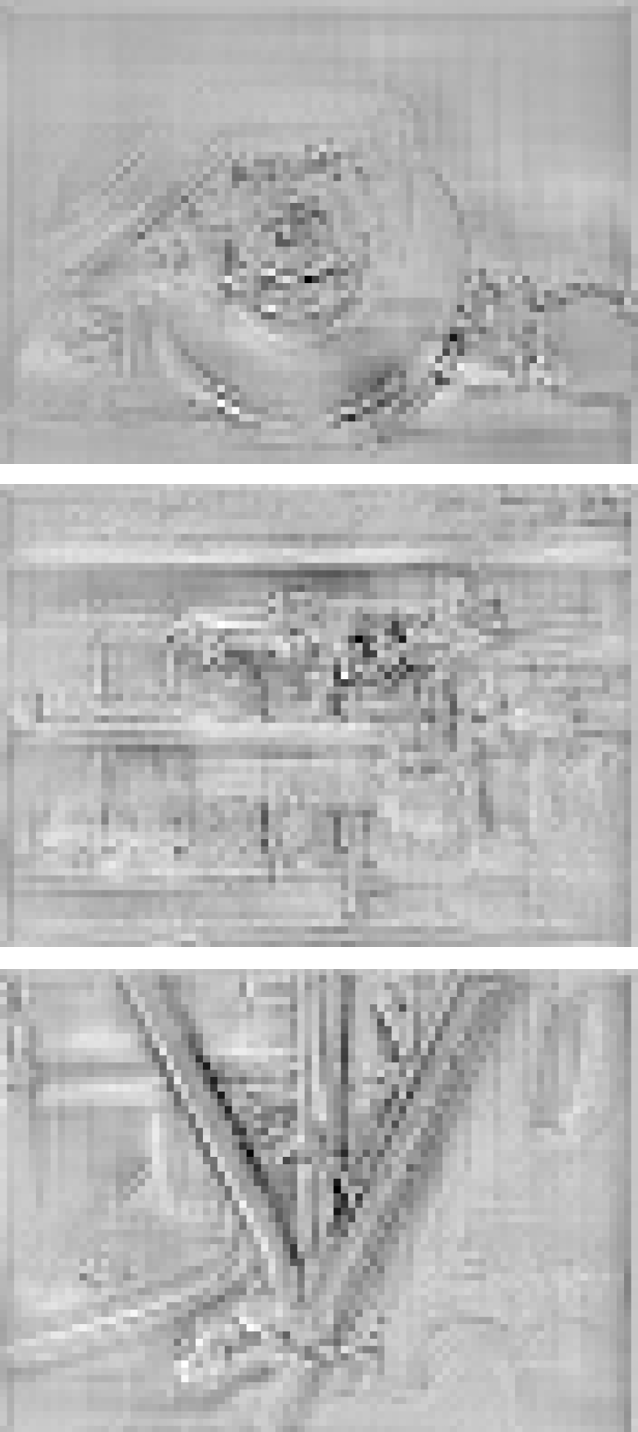}}
		\hspace{0.3em}
		\subfloat[]{\includegraphics[width=1.4in]{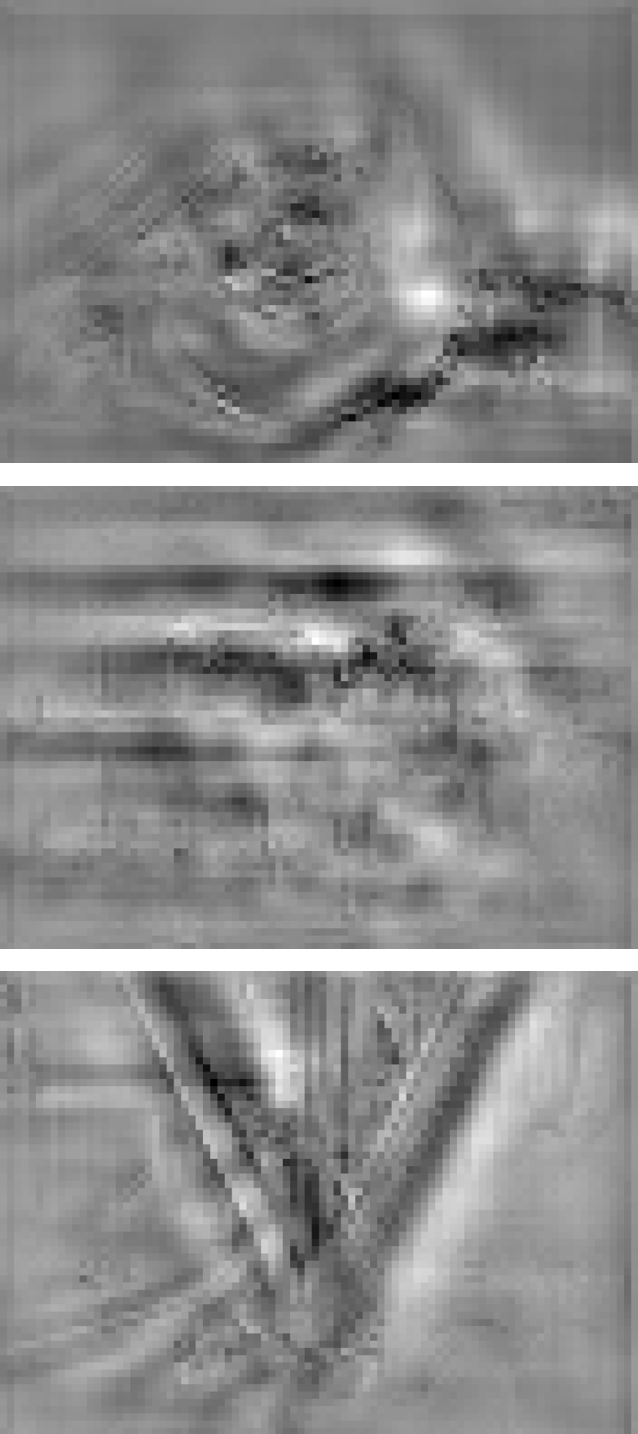}}
		\caption{Comparison of the average feature maps. (a) The feature maps of $F_{out}$ in DFB. (b) The feature maps output from DFB. The dilated convolution in the shortcut obtains the lost detail information through a larger receptive field.}
		\label{P_Dile}
	\end{center}
\end{figure}

Considering that the top layer has no additional information from the upper layer, we design a dilated fault bottleneck (DFB) to enrich the information as shown in Fig.~\ref{modules}(c). We add continuous dilated convolution to the shortcut, which is able to extend the receptive field under low computational costs and capture multi-scale context information~\cite{DeepLab}. Using DFB on the top layer can recover some of the detailed information lost during the previous processing.
Fig.~\ref{P_Dile}(a) is the feature map of $F_{out}$ in DFB, and Fig.~\ref{P_Dile}(b) is the feature map output from DFB. Through the larger receptive field obtained by the dilated convolution in the shortcut, we retrieve the detailed information and add it to the subsequent processing to improve the accuracy.

The maximum distance between two nonzero elements in dilated convolution is defined by hybrid dilated convolution:
\begin{equation}
	\label{111}
	L_i=\max \left[L_{i+1}-2r_i, 2r_i-L_{i+1},r_i \right].
\end{equation}
Among them, $L_i$ represents the maximum distance between two nonzero elements in the $i$-th layer, and $r_i$ represents the dilation rate of the $i$-th layer \cite{125}.
The convolution with a small dilation rate is concerned with short-distance information, while the convolution with a large dilation rate is concerned with long-distance information. With the help of dilated convolution, our method achieves a higher accuracy of faults on different scales.

\subsection{Loss Function}
\label{Loss Function}
As mentioned above, redundant and approximately repeated predictions are produced for each object during the inference process, which are removed by NMS in the post-processing. However, the computational costs are sharply increased by calculating the IoU of two overlapping boxes when using NMS. We make the network produce one-to-one predictions by calculating the classification loss function to eliminate the NMS, significantly reducing network calculations.
We use the detection head in our network like FCOS~\cite{FCOS}, but we do not calculate centerness loss because the definition of centerness leads to unexpected small ground-truth labels, which makes a possible set of ground-truth bounding boxes extremely hard to be recalled~\cite{Generalized_focal_loss}. Our loss function is defined as follows:
\begin{equation}
	\label{222}
	Loss=\lambda_{cls}\times L_{cls} + \lambda_{L1}\times L_{L1} + \lambda_{GIoU}\times L_{GIoU},
\end{equation}
where $L_{cls}$ is the classification loss of predicted classifications and ground-truth category label. $L_{L1}$ and $L_{GIoU}$ are samples $L1$ loss and GIoU loss~\cite{GiOU} between sample box and ground-truth box, respectively. $\lambda_{cls}$, $\lambda_{L1}$ and $\lambda_{GIoU}$ are coefficients of each component, and set as 2, 5 and 2, respectively.

Furthermore,  we use focal loss~\cite{Focal_Loss} in our $L_{cls}$:
\begin{equation}
	\label{focal_loss}
	{L_{cls}} = \left\{ {\begin{array}{*{20}{c}}
		{ - \alpha {{(1 - y)}^\beta }\log y\begin{array}{*{20}{c}}
		,&{x = 1}
		\end{array}}\\
		{ - (1 - \alpha ){y^\beta }\log (1 - y)\begin{array}{*{20}{c}}
		,&{x = 0}
		\end{array}}
		\end{array}} \right.,
\end{equation}
where $x$ is the label of the ground-truth, $y$ is the predicted fault, $\alpha$ and $\beta$ are balance factors set as 0.25 and 2, respectively. The $L1$ loss is defined as:
\begin{equation}
	\label{L1_loss}
	{L_{L1}} = \sum\limits_{i = 1}^n {\left| {{y_{true}} - {y_{predicted}}} \right|} .
\end{equation}
The $L1$ loss calculates the sum of the absolute value of the difference between all ground-truth values and predicted values. $GIoU$ loss is an improvement of IoU loss~\cite{IoU}, which is defined as:
\begin{equation}
	\label{GiOU}
	{L_{GIoU}} = 1 - IoU + \frac{{{A^c} - U}}{{{A^c}}} .
\end{equation}
The $GIoU$ loss not only pays attention to overlapping areas but also other non-overlapping areas, which can better reflect the overlap of the two boxes. $A^c$ represents the area of the smallest bounding rectangle between the prediction box and the ground-truth, and $U$ represents the union area between the prediction box and the ground-truth.
\section{EXPERIMENTS}
\setlength{\abovecaptionskip}{-0.2cm}  
\setlength{\belowcaptionskip}{0cm} 
\label{EXPERIMENTS}
\begin{table*}[t]
	\scriptsize
	\renewcommand\arraystretch{1.4}
	\caption{Comparison of different backbones.}
	\label{Overall network structure}
	\begin{center}
	\begin{tabular}{lp{12mm}<{\centering}p{12mm}<{\centering}p{12mm}<{\centering}p{12mm}<{\centering}p{12mm}<{\centering}p{12mm}<{\centering}p{12mm}<{\centering}p{12mm}<{\centering}}
	\toprule
	\multirow{2}{*}{Backbone} &
	\multirow{2}{*}{\begin{tabular}[c]{@{}c@{}}CDR$\uparrow$\\      (\%)\end{tabular}} &
	\multirow{2}{*}{\begin{tabular}[c]{@{}c@{}}FDR$\downarrow$\\      (\%)\end{tabular}} &
	\multirow{2}{*}{\begin{tabular}[c]{@{}c@{}}MDR$\downarrow$\\      (\%)\end{tabular}} &
	  \multicolumn{2}{c}{Time(s/image)} &
	  \multicolumn{2}{c}{Memory(MB)} &
	  \multirow{2}{*}{\begin{tabular}[c]{@{}c@{}}Model\\      size(MB)\end{tabular}} \\ \cmidrule(r){5-6} \cmidrule(r){7-8}
	&     &     &       & Train  & Test   & Train & Test &     \\ \midrule
	CSPDarkNet-53 \cite{CSPDarkNet-53}      & 97.79 & 1.90  & 0.31 & 0.185 & 0.020 & 9072  & 9292 & 446 \\
	ResNet18 \cite{ResNet}             & 95.03 & 0.07  & 4.90 & 0.253 & 0.006 & 1375  & 1051 & 163 \\
	ResNet34 \cite{ResNet}             & 90.33 & 0.97  & 8.70 & 0.100 & 0.009 & 2292  & 2333 & 279 \\
	ResNet50 \cite{ResNet}         & 91.44 & 0.07  & 8.49  & 0.198 & 0.014 & 8797  & 1147 & 355 \\
	ResNet101 \cite{ResNet}         & 91.23 & 0.03  & 8.73  & 0.241 & 0.021 & 6954  & 5092 & 566 \\
	SqueezeNet-V1.1 \cite{SqueezeNet}   & 62.69 & 27.93 & 9.39 & 0.098 & 0.009 & 2975  & 3036 & 46  \\
	DLA34 \cite{DLA}             & 96.72 & 2.66  & 0.62 & 0.115 & 0.010 & 2859  & 2914 & 212 \\
	EfficientNet-B0 \cite{EfficientNet-B0}    & 80.57 & 19.40 & 0.03 & 0.120 & 0.012 & 4171  & 4256 & 55  \\
	GhostNet \cite{GhostNet}          & 76.60 & 23.40 & 0.00 & 0.088 & 0.011 & 2008  & 2049 & 55  \\
	MnasNet-0.5 \cite{MnasNet}       & 72.01 & 27.99 & 0.00 & 0.055 & 0.007 & 1256  & 1283 & 23  \\
	PeleeNet \cite{Pelee}          & 86.50 & 12.22 & 1.28 & 0.100 & 0.013 & 1946  & 1977 & 66  \\
	RegNet-200M \cite{RegNet}        & 91.61 & 7.63  & 0.76 & 0.059 & 0.009 & 983   & 1006 & 40  \\
	VoVNet27-slim \cite{VoVNet27-slim}      & 95.58 & 2.31  & 2.11 & 0.108 & 0.010 & 3066  & 3129 & 119 \\
	MobileNetV3-Small \cite{MoilbNetv3Small}  & 81.22 & 18.71 & 0.07 & 0.047 & 0.007 & 709   & 724  & \textbf{21}  \\
	EfficientNetV2S \cite{EfficientNet-B0}   & 91.61 & 8.08  & 0.31 & 0.198 & 0.016 & 7732  & 7872 & 273 \\
	VGG-16 \cite{VGG}   & 74.63 & 17.81  & 7.56 & 0.124 & 0.023 & 7303  & 1645 & 675 \\
	InceptionNet-V3 \cite{InceptionV3}   & 95.17 & 2.11  & 2.73 & 0.094 & 0.009 & 5215  & 1069 & 203 \\ \bottomrule
	\end{tabular}
	\end{center}
	\end{table*}
In this section, we first introduce three fault datasets and evaluation metrics to verify the performance of NR FTI-FDet in the RVBS-FD of freight train.
Then we analyze the overall network, the effectiveness of FFP, and different modules in FFP. Finally, we compare our NR FTI-FDet with state-of-the-art methods.



\subsection{Experimental Setup}
\subsubsection{Implementation Details}
Our method is trained via backpropagation and the AdamW~\cite{AdamW} optimizer. The hyperparameters settings are the same for a fair comparison in all experiments. We train with a batch size of 16 and learning rate $5e^{-5}$ for 80K iterations, with learning rate dropped $10 \times$ at 60K iterations (following \cite{OneNet}). All experiments are performed in a PC environment with an Intel Core i9-9900K 3.60 GHz processor, 64G RAM, and a single GTX2080Ti GPU.
It is worth noting that we fine-tune the hyperparameters of each model to make the best performance on our hardware devices in Table~\ref{SOTA}.
\subsubsection{Datasets}
\label{Databases and Evaluation Metrics}
We use the labeling tool to create three datasets for the experiments \cite{Zhang_TIM}. The fault locations in the three data sets are located at block keys, dust collectors, and fastening bolts on the brake beam. These three parts are the critical components of the braking systems of freight trains to ensure transportation safety.

\begin{itemize}
	\item{\textbf{Bogie Block Key}: It is an important part to prevent the separation of the wheelset and the bogie. To detect the fault of this small part, we use 5440 images for training and 2897 images for testing, respectively.	}
	\item{\textbf{Dust Collector}: It is mainly used to filter impurities in the air and is usually installed next to the cut out cock. We use 815 images and 850 images for training and testing, respectively.}
	\item{\textbf{Fastening Bolt on Brake Beam}: It is an essential part of the train brake system to fix the brake beam. The horizontal force generated by train brakes is likely to cause the fastening bolt on brake beam to fall off or break. We use 1724 images and 1902 images for training and testing, respectively.}
\end{itemize}

\begin{table*}[t]
	\setlength{\abovecaptionskip}{-0.2cm}  
	\setlength{\belowcaptionskip}{0cm} 
	\scriptsize
	\renewcommand\arraystretch{1.4}
	\caption{The effectiveness of the proposed FEA, FBM and DFB in fault feature pyramid. The first row indicates that only FPN is added to the network, and the remaining rows suggest that the corresponding module is added to the FPN.}
	\label{Fault Feature Pyramid}
	\begin{center}
	\begin{tabular}{p{5mm}<{\centering}p{5mm}<{\centering}p{5mm}<{\centering}p{12mm}<{\centering}p{12mm}<{\centering}p{12mm}<{\centering}p{12mm}<{\centering}p{12mm}<{\centering}p{12mm}<{\centering}p{12mm}<{\centering}p{12mm}<{\centering}}
	\toprule
	\multirow{2}{*}{FEA} &
  \multirow{2}{*}{FBM} &
  \multirow{2}{*}{DFB} &
  \multirow{2}{*}{\begin{tabular}[c]{@{}c@{}}CDR$\uparrow$\\      (\%)\end{tabular}} &
  \multirow{2}{*}{\begin{tabular}[c]{@{}c@{}}FDR$\downarrow$\\      (\%)\end{tabular}} &
  \multirow{2}{*}{\begin{tabular}[c]{@{}c@{}}MDR$\downarrow$\\      (\%)\end{tabular}} &
  \multicolumn{2}{c}{Time(s/image)} &
  \multicolumn{2}{c}{Memory(MB)} &
  \multirow{2}{*}{\begin{tabular}[c]{@{}c@{}}Model\\      size(MB)\end{tabular}} \\ \cmidrule(r){7-8} \cmidrule(r){9-10}
  &   &   &       &      &      & Train  & Test   & Train & Test &      \\ \midrule
  --& --  &--   & 96.06  & 1.93& 2.00 & 0.321  & 0.011 & 6472  & 983  & 93.4 \\
\checkmark &--   & --  & 96.51  & 2.49 & 1.00& 0.316 & 0.011 & 6679  & 983  & 93.7 \\
--& \checkmark &  -- & 97.62  & 1.59 & 0.79& 0.309 & 0.011 & 6750  & 983  & 73.5 \\
--&  -- & \checkmark & 94.82  & 1.69 & 3.49& 0.414 & 0.013 & 7360  & 1013 & 117  \\
\checkmark & \checkmark & --  & 98.27  & 1.42 & 0.31& 0.307 & 0.011 & 6954  & 983  & 73.8 \\
\checkmark & --  & \checkmark & 98.07  & 1.76& 0.17& 0.412 & 0.014 & 7565  & 877  & 118  \\ \bottomrule
	\end{tabular}
	\end{center}
	\end{table*}

The above-mentioned typical faults are located at the bottom or on both sides of the train. However, there are various components around these parts, which bring great difficulties to fault detection. Meanwhile, due to the complex environment, the testing parts may be blocked by foreign objects, which also increases the difficulty of inspection. 

\subsubsection{Evaluation Metrics}
We use eight indexes to verify the effectiveness of the proposed NR FTI-FDet in the RVBS-FD of freight trains. They are false detection rate (FDR), correct detection rate (CDR), missed detection rate (MDR), training speed, testing speed, training memory usage, testing memory usage, and model size. Among them, FDR, CDR, and MDR are used to evaluate the accuracy of the detectors. Those three metrics are defined as follows: after using a method on the testing set with $m$ fault images and $n$ non-fault images, there are $a$ images considered the fault images, and $c$ images are detected as normal images. However, there are $b$ images in $a$ and $d$ images in $c$ are detected incorrectly, respectively, then:  $FDR = b/(m+n), MDR = d/(m+n),CDR = 1-FDR-MDR.$
The training and testing speed reflect the computational costs of the model. When the computing power is limited, we prefer to use models with low computational costs and fast detection speed. The memory usage of a model during training and testing reflects the dependence of a detector on the hardware system. These two metrics are also essential for fault detection tasks. 
We calculate the mean value of the FDR, CDR, and MDR as mFDR, mCDR, and mMDR to evaluate the different detectors on three fault datasets.

\subsection{Performance Analysis} 
\subsubsection{Backbone}
\label{Overall}
\begin{table*}[t]
	\setlength{\abovecaptionskip}{-0.2cm}  
	\setlength{\belowcaptionskip}{0cm} 
	\scriptsize
	\renewcommand\arraystretch{1.4}
	\caption{Comparison of different dilation rates in dilated fault bottleneck (DFB).}
	\label{Dilation Rate}
	\begin{center}
	\begin{tabular}{p{20mm}<{\centering}p{12mm}<{\centering}p{12mm}<{\centering}p{12mm}<{\centering}p{12mm}<{\centering}p{12mm}<{\centering}p{12mm}<{\centering}p{12mm}<{\centering}p{12mm}<{\centering}}
	\toprule
	\multirow{2}{*}{\begin{tabular}[c]{@{}c@{}}Dilation\\      Rate\end{tabular}} &
	\multirow{2}{*}{\begin{tabular}[c]{@{}c@{}}CDR$\uparrow$\\      (\%)\end{tabular}} &
  	\multirow{2}{*}{\begin{tabular}[c]{@{}c@{}}FDR$\downarrow$\\      (\%)\end{tabular}} &
  	\multirow{2}{*}{\begin{tabular}[c]{@{}c@{}}MDR$\downarrow$\\      (\%)\end{tabular}} &
	\multicolumn{2}{c}{Time(s/image)} &
	\multicolumn{2}{c}{Memory(MB)} &
	\multirow{2}{*}{\begin{tabular}[c]{@{}c@{}}Model\\      size(MB)\end{tabular}} \\ \cmidrule(r){5-6} \cmidrule(r){7-8}
	&   &    &      & Train  & Test   & Train & Test &     \\ \midrule
	{[}1, 2{]}       & 97.65 & 2.07 & 0.28 & 0.366 & 0.013 & 7354  & 1031 & 113 \\
	{[}1, 2, 5{]}     & 98.07 & 1.76 & 0.17 & 0.412 &0.014 & 7565  & 1033  & 118 \\
	{[}1, 2, 5, 1, 2{]} & 97.34 & 2.45 & 0.21 & 0.472 & 0.015  & 7979  & 1033 & 118 \\ \bottomrule
	\end{tabular}
	\end{center}
	\end{table*}
To verify the effectiveness of our NR FTI-FDet, we first explore the overall structure of the network. We replace the baseline network with various backbones for experiments, taking the bogie block key dataset as an example. As shown in Table~\ref{Overall network structure}, CSPDarkNet53 has the highest accuracy, but its model size is 446 MB, which is the largest model size among all tested models. The results show that the computational costs of CSPDrakNet53 are the largest, which can not meet our requirements because we hope the computational costs of the model can be sufficiently small. The model size of MobileNetV3-Small is the smallest, only 21 MB. Although the CDR is 81.22\%, it is more acceptable to us in terms of accuracy than MnasNet-0.5, which has almost the same model size. As for testing speed and memory usage, MobileNetV3-Small has the best performance among all test models. The results show that MobileNetV3-Small has a lower computational cost and is more suitable for resource-limited environments. Therefore, we choose MobileNetV3-Small as the backbone of our network.


\subsubsection{Fault Feature Pyramid}
\label{SFFP}
To improve the performance, we propose a traditional feature pyramid to construct feature maps in different sizes with advanced semantic information.
In Table~\ref{Fault Feature Pyramid}, when we add the traditional FPN to the network, the CDR is increased from 81.22\% to 96.0\%. However, the test time is $2 \times$ longer than the original network, and the training time is $7 \times$ longer.
The memory usage of this model is also increased. The larger memory usage of the model means that the model needs a better hardware environment to run normally, which is not conducive to model deployment in the wild. 
Meanwhile, the model size is growing from 21.0 MB to 93.4 MB. 

To verify the effectiveness of FEA, FEB, and DFB, we add these three modules to FPN in turn. As shown in Table~\ref{Fault Feature Pyramid}, with the addition of FEA, the model size is only increased by 0.3 MB and a 0.45\% increase in accuracy. The results show that FEA can improve accuracy with a slight increase in computational costs.
In addition, with FBM added to the network, the model size is reduced from 93.4 MB to 73.5 MB. The CDR is also increased from 96.06\% to 97.62\%, which is better than adding FEA alone. However, the addition of DFB makes a decrease in accuracy, and the model size is also increased to 117 MB. The effect of using DFB alone is not as good as using DFB with other modules.

When adding FEA and DFB to the network simultaneously, the effect is better than adding FEA or DFB separately. The model size only increases 1 MB compared with using DFB alone. Similarly, when FEA and FBM are added to the network simultaneously, the detection accuracy is further improved. Compared with adding FBM alone, the CDR is increased from 97.62\% to 98.27\%, and the model size is only increased by 0.3 MB. The training speed, testing speed, and memory usage are almost unchanged. Therefore, FEA and FBM can improve detection accuracy without increasing a large number of computational costs.

\subsubsection{Dilation Rates of DFB}
\label{Number of dilated convolutions}
Dilated convolutions with different dilation rates are stacked to obtain different sizes of receptive fields. In Table~\ref{Dilation Rate}, we conduct constant stacking experiments based on the dilated convolution with dilation rates of 1 and 2. 
The network uses MobileNetV3-Small as the backbone, and the fault feature pyramid uses a combination of FEA and DFB. When testing on the bogie block key dataset, we find stacking dilation rates of [1, 2, 5] have the best accuracy, while CDR achieves 98.07\%. The results are decreased when reducing the number of dilated convolutions or adding a group of dilated convolutions. The test memory usage using the dilation rates of [1, 2, 5] is the smallest when the model size is the same, which shows that the dilation rates of [1, 2, 5] are more suitable to detect the fault in train images.


\subsubsection{Distribution of FBM and DFB}
In Table~\ref{Distribution of FBM and DFB}, the tested network uses MobileNetV3-Small as the backbone and adds FEA and FBM to FPN. The dilation rates are [1, 2, 5] for stacking and testing on the bogie block key dataset. We use FBM instead if DFB is not used in a specific layer. 
It can be seen from Table~\ref{Distribution of FBM and DFB} that the accuracy decreases when DFB is not used in all layers. Compared to the first and second row, the accuracy drops from 98.27\% to 95.24\% because the second row uses the DFB in the bottom layer. 
The bottom layer uses the dilated convolutions to obtain a larger receptive field, which leads to more invalid information and a decrease in accuracy. 
The accuracy is improved when not use DFB at the bottom layer and only use it at the top layer. We can conclude from Table~\ref{Distribution of FBM and DFB} that DFB is more suitable for the top layer. It helps the top layer obtain more detailed information, but using DFB in other layers causes information redundancy and affects accuracy. Therefore, the FBM is more suitable for other layers to accept the rich information better transmitted from the upper layer.

\begin{table}[t]
	\setlength{\abovecaptionskip}{-0.2cm}  
	\setlength{\belowcaptionskip}{0cm} 
	\scriptsize
	\renewcommand\arraystretch{1.4}
	\caption{The Comparison of FBM ($\circ$ ) and DFB ($\bullet $) in fault feature pyramid.}
	\label{Distribution of FBM and DFB}
	\begin{center}
	\begin{tabular}{p{5mm}<{\centering}p{5mm}<{\centering}p{5mm}<{\centering}cccc}
		\toprule
	Level3 & Level4 & Level5 & CDR(\%)$\uparrow$ & FDR(\%)$\downarrow$ & MDR(\%)$\downarrow$ & \begin{tabular}[c]{@{}c@{}}Model\\ Size(MB)\end{tabular} \\\midrule
	$\circ$    & $\circ$  & $\circ$  & 98.27 & 1.42 & 0.31 & 73.8 \\ 
	$\bullet $ & $\circ$  & $\circ$  & 95.24 & 2.35 & 2.42 & 95.9 \\
	$\circ$  & $\bullet $ & $\circ$  & 97.58 & 1.45 & 0.97 & 95.9 \\
	$\circ$  & $\circ$  & $\bullet $ & 98.79 & 1.04 & 0.17 & 95.9 \\
	$\circ$  & $\bullet $ & $\bullet $ & 92.47 & 2.80 & 4.73 & 118 \\
	$\bullet $ & $\circ$  & $\bullet $ & 97.55 & 2.14 & 0.31 & 118 \\
	$\bullet $ & $\bullet $ & $\circ$  & 97.00 & 2.14 & 0.86 & 118 \\
	$\bullet $ & $\bullet $ & $\bullet $ & 98.07 & 1.76 & 0.17 & 118 \\ \bottomrule
	\end{tabular}
	\end{center}
	\end{table}


\subsubsection{Loss Function}
\begin{table}[]
	\setlength{\abovecaptionskip}{-0.2cm}  
	\setlength{\belowcaptionskip}{0cm} 
	\scriptsize
	\renewcommand\arraystretch{1.4}
	\caption{Compare the importance of different loss functions and the test time consumed by NMS.}
	\label{loss}
	\begin{center}
	\begin{tabular}{cccccc}
		\toprule
	\multirow{2}{*}{$L_{cls}$} & \multirow{2}{*}{$L_{L1}$} & \multirow{2}{*}{$L_{GIoU}$} & \multirow{2}{*}{CDR(\%)$\uparrow$} & \multicolumn{2}{c}{Time(s/image)} \\ \cmidrule(r){5-6}
	  &   &   &          & Test & Test (+NMS) \\ \midrule
	  \checkmark& -- & -- & 58.06 & 0.012   & 0.032 (+0.021)  \\
	-- & \checkmark& -- & 60.82 & 0.012   & 0.025 (+0.013)  \\
	-- & -- & \checkmark & 59.65 & 0.012   & 0.024 (+0.012)  \\
	\checkmark & \checkmark & -- & 82.84 & 0.011   & 0.033 (+0.022)  \\
	-- & \checkmark & \checkmark & 60.86 & 0.013   & 0.025 (+0.012)  \\
	\checkmark & \checkmark & \checkmark & 98.79    & 0.012   & 0.027 (+0.015)  \\ \bottomrule
	\end{tabular}
\end{center}
	\end{table}
	
	To verify the importance of loss functions calculated in the head, we apply them to the training process separately, as shown in Table~\ref{loss}. The loss functions we used can be divided into two roles, where $L1$ loss and GIoU loss are used for object localization. The $L_{cls}$ is a classification loss that eliminates post-processing by generating a unique prediction. The accuracy is maintained at roughly 60\% when using $L1$ loss and GIou loss, or only one of these two loss functions. With the addition of $L_{cls}$, the accuracy is dramatically improved which further emphasizes the importance of classification loss for train fault detection. In addition, we included NMS in the test of each experiment to compare the test time consumed by NMS post-processing. The test time is $2\times$ longer after adding NMS, as seen in Table~\ref{loss}.

\subsection{Comparison with State-of-the-Art Methods}
To further illustrate the superiority of our NR FTI-FDet, we compare our framework with state-of-the-art methods. These detectors can be divided into NMS-based and NMS-free methods. It is worth noting that the parameters in each method are tuned to be the best performance on our datasets, and the memory usage during training represents the memory consumed by each batch.

\begin{table*}[t]
	\scriptsize
	\renewcommand\arraystretch{1.4}
	\caption{Compare with the state-of-the-art methods on three typical fault datasets.}
	\label{SOTA}
	\begin{center}
	\begin{tabular}{lcp{10mm}<{\centering}p{10mm}<{\centering}p{10mm}<{\centering}p{8mm}<{\centering}p{8mm}<{\centering}p{8mm}<{\centering}p{8mm}<{\centering}cc}
	\toprule
	\multirow{2}{*}{Methods} &
	  \multirow{2}{*}{Backbone} &
	  \multirow{2}{*}{\begin{tabular}[c]{@{}c@{}}mCDR$\uparrow$\\      (\%)\end{tabular}} &
	  \multirow{2}{*}{\begin{tabular}[c]{@{}c@{}}mFDR$\downarrow$\\      (\%)\end{tabular}} &
	  \multirow{2}{*}{\begin{tabular}[c]{@{}c@{}}mMDR$\downarrow$\\      (\%)\end{tabular}} &
	  \multirow{2}{*}{\begin{tabular}[c]{@{}c@{}}Batch\\ size\end{tabular}} &
	  \multicolumn{2}{c}{Time(s/image)} &
	  \multicolumn{2}{c}{Memory(MB)} &
	  \multirow{2}{*}{\begin{tabular}[c]{@{}c@{}}Model\\      size(MB)\end{tabular}}	   \\ \cmidrule(r){7-8} \cmidrule(r){9-10}
						 &                  &       &      &    &   & Train & Test  & Train** & Test &            \\ \midrule
	\multicolumn{11}{c}{\textbf{Traditional Detectors}}  \\ \midrule
	Cascade detector (LBP)\cite{OE} & --        & 92.65 & 4.95 & 2.40   & -- & -- & 0.048 & --  & -- & 0.12  \\ 
	HOG+Adaboost+SVM\cite{Zhang_TIM} & --       & 96.53 	&2.54 	&0.93   & -- & -- & 0.049 & --  & -- & 0.11  \\
	FAMRF+EHF\cite{OE} & --        & 95.51 	&3.58	&0.9033    & -- & -- & 0.725 & --  & -- & --  \\ \midrule
	\multicolumn{11}{c}{\textbf{NMS-based Two-stage Detectors}}  \\ \midrule
	Cascade R-CNN~\cite{Cascade_R-CNN}       & ResNet-50        & \textbf{99.46} & \textbf{0.35} & 0.19   & 6 & 1.195 & 0.061 & 1506  & 1841 & 552.6  \\
	Faster R-CNN \cite{Faster_R-CNN}         & ResNet-50        & 98.41 & 1.05 & 0.54  & 6 & 0.956 & 0.055 & 1453  & 1575 & 330.3   \\
	Dynamic R-CNN \cite{Dynamic_R-CNN}        & ResNet-50        & 99.30 & 0.44 & 0.26  & 2 & 0.424 & 0.056 & 1708  & 1427 & 330.2   \\
	Grid R-CNN \cite{Grid_R-CNN}           & ResNet-50        & 97.90 & 0.86 & 1.24  & 4 & 0.831 & 0.064 & 1627  & 1523 & 515.2   \\
	RepPoints \cite{RepPoints}            & ResNet-50   & 99.69 & 0.17 & 0.14  & 4 & 0.812 & 0.058 & 1277  & 1427 & 294.0   \\ 
	FTI-FDet * \cite{Zhaiyao_Zhang}    & VGG16        & 99.47 & 0.12 & 0.41  & 128 & -- & 0.071 & --  & 1823 & 557.3   \\
	Light FTI-FDet * \cite{Zhang_TIM}  & VGG16        & 99.79 & 0.18 & 0.03  & 128  & -- & 0.058 & --  & 1533 & 89.7  \\
	LR FTI-FDet * \cite{Zhang_TII}  & RFDNet \cite{Zhang_TII}       & 99.62 & 0.12 & 0.26  & 256  & 0.135 & 0.026 & --  & 713 & 19.6  \\ \midrule
	\multicolumn{11}{c}{\textbf{NMS-based One-stage Detectors}}    \\ \midrule
	RetinaNet \cite{Focal_Loss}	     	 & ResNet-50        & 89.21 & 2.25 & 8.54  & 6  & 0.828 & 0.051 & 1123  & 1573 & 290.2  \\
	YOLOv3 \cite{YOLOv3}               	& DarkNet53        & 92.87 & 3.66 & 3.47  & 8 & 0.757 & 0.018 & 896   & 1321 & 492.5   \\
	CentripetalNet \cite{CentripetalNet}  & HourglassNet-104 & 97.15 & 1.00 & 1.85  & 2 & 1.252 & 0.297 & 3878  & 2901 & 2469.1  \\
	CornerNet \cite{CornerNet}            & HourglassNet-104 & 98.13 & 1.17 & 0.70  & 2 & 0.804 & 0.245 & 3643  & 2337 & 2412.4  \\
	FCOS \cite{FCOS}               		 & ResNet-50        & 99.45 & 0.40 & 0.15  & 6 & 0.835 & 0.050 & 1311  & 1283  & 256.1   \\
	GFL \cite{Generalized_focal_loss}     & ResNet-101       & 98.59 & 1.34 & 0.07  & 4 & 0.848 & 0.070 & 2217  & 1359 & 409.7   \\
	SSD \cite{SSD}                  		& VGG16            & 93.13 & 0.76 & 6.11 & 16 & 0.915 & 0.021  & 639   & 1157 & 190.0  \\
	VarifocalNet \cite{VarifocalNet}      & ResNet-50        & 96.98 & 0.60 & 2.42  & 4 & 1.239 & 0.060  & 1454  & 1551 & 261.1   \\
	YOLOF \cite{YOLOF}                & ResNet-50        & 92.89 & 5.68 & 1.43  & 8 & 0.649 & 0.036 & 893   & 1321 & 338.0   \\
	YOLOX \cite{YOLOX}              & Modified CSP v5  & 97.17 & 2.81 & \textbf{0.02}  & 2 & 0.259 & 0.026 & 1932  & 1265 & 650.7   \\ \midrule
	\multicolumn{11}{c}{\textbf{NMS-free Detectors}}      \\ \midrule
	Libra R-CNN \cite{Balanced}          & ResNet-50        & 97.77 & 0.88 & 1.35  & 6 & 1.117 & 0.063 & 1614  & 1573 & 332.4   \\
	CenterNet \cite{Object}            & ResNet-18        & 95.13 & 4.41 & 0.46  & 16 & 0.352 & 0.013 & \textbf{162}   & 1065 & 113.8  \\
	Deformable DETR \cite{Deformable_DETR}      & ResNet-50   & 84.26 & 13.65 & 2.09  & 2 & 0.314 & 0.043 & 2009  & 1229 & 482.1   \\
	Sparse R-CNN \cite{Sparse_R-CNN}         & ResNet-50        & 87.56 & 12.40 & 0.04 & 6  & 1.008 & 0.058 & 1538  & 1615 & 1272.6 \\ 
	OneNet \cite{OneNet}         & ResNet-50        & 82.14 & 14.19 & 3.67 & 4 & \textbf{0.200} & 0.014 & 2199  & 1147 & 355.0  \\\midrule
	NR FTI-FDet                & MobileNetV3-Small & 98.40 & 0.99 & 0.61  & 16  & 0.316 & \textbf{0.012} & 436   & \textbf{1011} & \textbf{95.9} \\ \bottomrule
	\end{tabular}
	\begin{tablenotes}
		*The method was experimented with using the publicly available Caffe~\cite{caffe} on a single GTX1080Ti.\\
		**Relevant parameters in each detector are fine-tuned for optimal performance, but the batch size directly affects the memory usage during training. For a fair comparison, the memory usage during training in the table represents the memory consumed by each batch.
	  \end{tablenotes}
	\end{center}
\end{table*}

\subsubsection{Comparison of Accuracy and Model Size}
In Table~\ref{SOTA}, the mCDR of NR FTI-FDet reaches 98.40\%, which is higher than all NMS-free detectors, almost two-stage detectors, and one-stage detectors. The mCDR of Light FTI-FDet is higher than our method by 1.39\%. But this detection speed is $5\times$ larger than our NR FTI-FDet.
As for model size, our NR FTI-FDet has the smallest model size among all methods except Light FTI-FDet and LR FTI-FDet. The model size of CenterNet is 113.8 MB which uses ResNet18 as the backbone, but mCDR of CenterNet is still 3.27\% lower than our method. The results show that using a lightweight backbone can affect the final model size to a certain extent. And our optimization for FPN with DFB and FBM also further reduces the model size. 
The NMS-based detectors have a higher mCDR but a larger model size. It is not worthwhile to trade a slight increase in accuracy for doubling the model size in a resource-limited environment.

\subsubsection{Comparison of Detection Speed and Memory Usage}

Computational cost is often reflected in speed, memory usage, and model size.
As for memory usage, our training memory usage is 436 MB. The training memory of one-stage NMS-based detectors occupies more than 1000 MB. The two-stage NMS-based detectors are generally larger because these methods have high computational complexity. It is more practical to use a one-stage detector when resources are limited.

In terms of training and testing speed, our method only takes 0.316 seconds to train one image, faster than all two-stage detectors in Table~\ref{SOTA}. Among the one-stage detectors, only YOLOX has a shorter training time than ours, OneNet and deformable DETR in the NMS-free methods are slightly faster than ours. Nevertheless, our method is the fastest in terms of testing speed, only taking 0.012 seconds to test one image (\textgreater 83 FPS), as shown in Table~\ref{SOTA}.
The above experimental results show that our NR FTI-FDet improves the detection speed at a slight sacrifice of accuracy and model size compared with our previous work. Moreover, our method has a lower demand for hardware resources, making it more suitable for deploying RVBS-FD of freight trains in the field.

\section{CONCLUSION}
\label{6}
This paper proposes a real-time, NMS-free, and high accuracy detector named NR FTI-FDet in the RVBS-FD of freight trains. Our detector consists of a lightweight backbone, fault feature pyramid and detection head. The fault feature pyramid is composed of fault enhance attention, dilated fault bottleneck, and fault bottleneck module. Experiments on three datasets show that the detection speed of our NR FTI-FDet in the RVBS-FD of freight trains is fastest than the state-of-the-art detectors, reaching over 83 FPS. Meanwhile, the model size of our detector is only 95.9 MB.
Finally, our RVBS-FD requires low hardware resources during training and testing, and the accuracy still reaches 98.4\%.

We will focus on optimizing our framework in the following three areas in the future. First, the accuracy of RVBS-FD should be further improved while ensuring the detection speed and model size.  Second, we plan to use different image processing methods during training and testing to increase detection speed. Third, our NR FTI-FDet can be deployed to edge computing equipment after further optimization.

\bibliographystyle{ieeetr}
\bibliography{zy}

\begin{thebibliography}{10}

\bibitem{Zhang_TIM}
Y.~Zhang, M.~Y. Liu, Y.~A. Chen, H.~J. Zhang, and Y.~W. Guo, ``{Real-Time
  Vision-Based System of Fault Detection for Freight Trains},'' {\em IEEE
  Transactions on Instrumentation and Measurement}, vol.~69, no.~7,
  pp.~5274--5284, 2020.

\bibitem{4}
L.~Liu, F.~Zhou, and Y.~He, ``{Automated Visual Inspection System for Bogie
  Block Key Under Complex Freight Train Environment},'' {\em IEEE Transactions
  on Instrumentation and Measurement}, vol.~65, no.~1, pp.~2--14, 2016.

\bibitem{OE}
G.~Sun, Y.~Zhang, H.~Tang, H.~Zhang, M.~Liu, and D.~Zhao, ``{Railway Equipment
  Detection Using Exact Height Function Shape Descriptor Based on Fast Adaptive
  Markov Random Field},'' {\em Optical Engineering}, vol.~57, no.~5, pp.~1 --
  14, 2018.

\bibitem{3}
S.~Lu, Z.~Liu, and Y.~Shen, ``{Automatic Fault Detection of Multiple Targets in
  Railway Maintenance Based on Time-Scale Normalization},'' {\em IEEE
  Transactions on Instrumentation and Measurement}, vol.~67, no.~4,
  pp.~849--865, 2018.

\bibitem{2}
R.~S. Pahwa, J.~Chao, J.~Paul, Y.~Li, M.~T. Lay~Nwe, S.~Xie, A.~James,
  A.~Ambikapathi, Z.~Zeng, and V.~R. Chandrasekhar, ``{FaultNet: Faulty
  Rail-Valves Detection using Deep Learning and Computer Vision},'' in {\em
  IEEE Intelligent Transportation Systems Conference}, pp.~559--566, 2019.

\bibitem{Fu}
X.~Fu, K.~L. Li, J.~Liu, K.~Q. Li, Z.~Zeng, and C.~Chen, ``{A two-stage
  attention aware method for train bearing shed oil inspection based on
  convolutional neural networks},'' {\em Neurocomputing}, vol.~380,
  pp.~212--224, 2020.

\bibitem{Sun}
J.~H. Sun, Z.~W. Xiao, and Y.~X. Xie, ``{Automatic multi-fault recognition in
  TFDS based on convolutional neural network},'' {\em Neurocomputing},
  vol.~222, pp.~127--136, 2017.

\bibitem{7}
L.~Chang, Z.~Liu, Y.~Shen, and G.~Zhang, ``{Novel Multistate Fault Diagnosis
  and Location Method for Key Components of High-Speed Trains},'' {\em IEEE
  Transactions on Industrial Electronics}, vol.~68, no.~4, pp.~3537--3547,
  2021.

\bibitem{Ye}
T.~Ye, Z.~H. Zhang, X.~Zhang, Y.~R. Chen, and F.~Q. Zhou, ``{Fault detection of
  railway freight cars mechanical components based on multi-feature fusion
  convolutional neural network},'' {\em International Journal of Machine
  Learning and Cybernetics}, vol.~12, no.~6, pp.~1789--1801, 2021.

\bibitem{XiaoL}
L.~Xiao, B.~Wu, and Y.~Hu, ``{Missing Small Fastener Detection Using Deep
  Learning},'' {\em IEEE Transactions on Instrumentation and Measurement},
  vol.~70, pp.~1--9, 2021.

\bibitem{5}
C.~Chen, X.~Zou, Z.~Zeng, Z.~Cheng, L.~Zhang, and S.~C.~H. Hoi, ``{Exploring
  Structural Knowledge for Automated Visual Inspection of Moving Trains},''
  {\em IEEE Transactions on Cybernetics}, pp.~1--14, 2020.

\bibitem{OneNet}
P.~Sun, Y.~Jiang, E.~Xie, W.~Shao, Z.~Yuan, C.~Wang, and P.~Luo, ``{What Makes
  for End-to-End Object Detection},'' in {\em International Conference on
  Machine Learning}, pp.~9934--9944, 2021.

\bibitem{Sun.J}
J.~H. Sun, Y.~X. Xie, and X.~Q. Cheng, ``{A Fast Bolt-Loosening Detection
  Method of Running Train's Key Components Based on Binocular Vision},'' {\em
  IEEE Access}, vol.~7, pp.~32227--32239, 2019.

\bibitem{Ling}
L.~Xiao, B.~Wu, Y.~M. Hu, and J.~Liu, ``{A Hierarchical Features-Based Model
  for Freight Train Defect Inspection},'' {\em IEEE Sensors Journal}, vol.~20,
  no.~5, pp.~2671--2678, 2020.

\bibitem{MoilbNetv3Small}
A.~Howard, M.~Sandler, B.~Chen, W.~Wang, L.-C. Chen, M.~Tan, G.~Chu,
  V.~Vasudevan, Y.~Zhu, R.~Pang, H.~Adam, and Q.~Le, ``Searching for
  mobilenetv3,'' in {\em IEEE/CVF International Conference on Computer Vision},
  pp.~1314--1324, 2019.

\bibitem{Pelee}
R.~J. Wang, X.~Li, and C.~X. Ling, ``{Pelee: A Real-Time Object Detection
  System on Mobile Devices},'' in {\em The 32nd International Conference on
  Neural Information Processing Systems}, pp.~1967--1976, 2018.

\bibitem{MnasNet}
M.~Tan, B.~Chen, R.~Pang, V.~Vasudevan, M.~Sandler, A.~Howard, and Q.~V. Le,
  ``{MnasNet: Platform-Aware Neural Architecture Search for Mobile},'' in {\em
  IEEE/CVF Conference on Computer Vision and Pattern Recognition},
  pp.~2815--2823, 2019.

\bibitem{GhostNet}
K.~Han, Y.~Wang, Q.~Tian, J.~Guo, C.~Xu, and C.~Xu, ``{GhostNet: More Features
  From Cheap Operations},'' in {\em IEEE/CVF Conference on Computer Vision and
  Pattern Recognition}, pp.~1577--1586, 2020.

\bibitem{RegNet}
I.~Radosavovic, R.~P. Kosaraju, R.~Girshick, K.~He, and P.~Dollár,
  ``{Designing Network Design Spaces},'' in {\em IEEE/CVF Conference on
  Computer Vision and Pattern Recognition}, pp.~10425--10433, 2020.

\bibitem{SqueezeNet}
F.~N. {Iandola}, S.~{Han}, M.~W. {Moskewicz}, K.~{Ashraf}, W.~J. {Dally}, and
  K.~{Keutzer}, ``{{SqueezeNet: AlexNet-level accuracy with 50x fewer
  parameters and 0.5MB model size}},'' {\em arXiv e-prints arXiv:1602.07360},
  2016.

\bibitem{EfficientNet-B0}
M.~Tan and Q.~Le, ``{EfficientNet: Rethinking Model Scaling for Convolutional
  Neural Networks},'' in {\em Proceedings of the 36th International Conference
  on Machine Learning}, pp.~6105--6114, 2019.

\bibitem{ResNet}
K.~He, X.~Zhang, S.~Ren, and J.~Sun, ``{Deep Residual Learning for Image
  Recognition},'' in {\em IEEE Conference on Computer Vision and Pattern
  Recognition}, pp.~770--778, 2016.

\bibitem{CSPDarkNet-53}
C.-Y. Wang, H.-Y. Mark~Liao, Y.-H. Wu, P.-Y. Chen, J.-W. Hsieh, and I.-H. Yeh,
  ``{CSPNet: A New Backbone that can Enhance Learning Capability of CNN},'' in
  {\em IEEE/CVF Conference on Computer Vision and Pattern Recognition
  Workshops}, pp.~1571--1580, 2020.

\bibitem{DLA}
F.~Yu, D.~Wang, E.~Shelhamer, and T.~Darrell, ``{Deep Layer Aggregation},'' in
  {\em IEEE/CVF Conference on Computer Vision and Pattern Recognition},
  pp.~2403--2412, 2018.

\bibitem{VoVNet27-slim}
Y.~Lee, J.~Hwang, S.~Lee, Y.~Bae, and J.~Park, ``{An Energy and GPU-Computation
  Efficient Backbone Network for Real-Time Object Detection},'' in {\em
  IEEE/CVF Conference on Computer Vision and Pattern Recognition Workshops},
  pp.~752--760, 2019.

\bibitem{Faster_R-CNN}
S.~Ren, K.~He, R.~Girshick, and J.~Sun, ``{Faster R-CNN: Towards Real-Time
  Object Detection with Region Proposal Networks},'' {\em IEEE Transactions on
  Pattern Analysis and Machine Intelligence}, vol.~39, no.~6, pp.~1137--1149,
  2017.

\bibitem{Grid_R-CNN}
X.~Lu, B.~Li, Y.~Yue, Q.~Li, and J.~Yan, ``{Grid R-CNN},'' in {\em IEEE/CVF
  Conference on Computer Vision and Pattern Recognition}, pp.~7355--7364, 2019.

\bibitem{RepPoints}
Z.~Yang, S.~Liu, H.~Hu, L.~Wang, and S.~Lin, ``{RepPoints: Point Set
  Representation for Object Detection},'' in {\em IEEE/CVF International
  Conference on Computer Vision}, pp.~9656--9665, 2019.

\bibitem{Dynamic_R-CNN}
H.~Zhang, H.~Chang, B.~Ma, N.~Wang, and X.~Chen, ``{Dynamic R-CNN: Towards High
  Quality Object Detection via Dynamic Training},'' in {\em European Conference
  on Computer Vision}, pp.~260--275, 2020.

\bibitem{Cascade_R-CNN}
Z.~Cai and N.~Vasconcelos, ``{Cascade R-CNN: High Quality Object Detection and
  Instance Segmentation},'' {\em IEEE Transactions on Pattern Analysis and
  Machine Intelligence}, vol.~43, no.~5, pp.~1483--1498, 2021.

\bibitem{SSD}
W.~Liu, D.~Anguelov, D.~Erhan, C.~Szegedy, S.~Reed, C.-Y. Fu, and A.~C. Berg,
  ``{SSD: Single Shot MultiBox Detector},'' in {\em European Conference on
  Computer Vision}, pp.~21--37, 2016.

\bibitem{YOLOv3}
J.~{Redmon} and A.~{Farhadi}, ``{YOLOv3: An Incremental Improvement},'' {\em
  arXiv e-prints arXiv:1804.02767}, 2018.

\bibitem{YOLOF}
Q.~Chen, Y.~Wang, T.~Yang, X.~Zhang, J.~Cheng, and J.~Sun, ``{You Only Look
  One-level Feature},'' in {\em IEEE/CVF Conference on Computer Vision and
  Pattern Recognition}, pp.~13034--13043, 2021.

\bibitem{YOLOX}
Z.~{Ge}, S.~{Liu}, F.~{Wang}, Z.~{Li}, and J.~{Sun}, ``{YOLOX: Exceeding YOLO
  Series in 2021},'' {\em arXiv e-prints arXiv:2107.08430}, 2021.

\bibitem{CornerNet}
H.~Law and J.~Deng, ``{CornerNet: Detecting Objects as Paired Keypoints},'' in
  {\em European Conference on Computer Vision}, pp.~765--781, 2018.

\bibitem{FCOS}
Z.~Tian, C.~Shen, H.~Chen, and T.~He, ``{FCOS: Fully Convolutional One-Stage
  Object Detection},'' in {\em IEEE/CVF International Conference on Computer
  Vision}, pp.~9626--9635, 2019.

\bibitem{CentripetalNet}
Z.~Dong, G.~Li, Y.~Liao, F.~Wang, P.~Ren, and C.~Qian, ``{CentripetalNet:
  Pursuing High-Quality Keypoint Pairs for Object Detection},'' in {\em
  IEEE/CVF Conference on Computer Vision and Pattern Recognition},
  pp.~10516--10525, 2020.

\bibitem{Generalized_focal_loss}
X.~Li, W.~Wang, L.~Wu, S.~Chen, X.~Hu, J.~Li, J.~Tang, and J.~Yang,
  ``{Generalized Focal Loss: Learning Qualified and Distributed Bounding Boxes
  for Dense Object Detection},'' in {\em Advances in Neural Information
  Processing Systems}, vol.~33, pp.~21002--21012, 2020.

\bibitem{Focal_Loss}
T.~Lin, P.~Goyal, R.~Girshick, K.~He, and P.~Dollár, ``{Focal Loss for Dense
  Object Detection},'' {\em IEEE Transactions on Pattern Analysis and Machine
  Intelligence}, vol.~42, no.~2, pp.~318--327, 2020.

\bibitem{VarifocalNet}
H.~Zhang, Y.~Wang, F.~Dayoub, and N.~Sünderhauf, ``{VarifocalNet: An IoU-aware
  Dense Object Detector},'' in {\em IEEE/CVF Conference on Computer Vision and
  Pattern Recognition}, pp.~8510--8519, 2021.

\bibitem{Object}
X.~{Zhou}, D.~{Wang}, and P.~{Kr{\"a}henb{\"u}hl}, ``{Objects as Points},''
  {\em arXiv e-prints arXiv:1904.07850}, 2019.

\bibitem{Sparse_R-CNN}
P.~Sun, R.~Zhang, Y.~Jiang, T.~Kong, C.~Xu, W.~Zhan, M.~Tomizuka, L.~Li,
  Z.~Yuan, C.~Wang, and P.~Luo, ``{Sparse R-CNN: End-to-End Object Detection
  with Learnable Proposals},'' in {\em IEEE/CVF Conference on Computer Vision
  and Pattern Recognition}, pp.~14449--14458, 2021.

\bibitem{Deformable_DETR}
X.~{Zhu}, W.~{Su}, L.~{Lu}, B.~{Li}, X.~{Wang}, and J.~{Dai}, ``{Deformable
  DETR: Deformable Transformers for End-to-End Object Detection},'' in {\em
  International Conference on Learning Representations}, 2021.

\bibitem{Balanced}
J.~Pang, K.~Chen, Q.~Li, Z.~Xu, H.~Feng, J.~Shi, W.~Ouyang, and D.~Lin,
  ``{Towards Balanced Learning for Instance Recognition},'' {\em International
  Journal of Computer Vision}, vol.~129, no.~5, pp.~1376--1393, 2021.

\bibitem{FPN}
T.-Y. Lin, P.~Dollár, R.~Girshick, K.~He, B.~Hariharan, and S.~Belongie,
  ``{Feature Pyramid Networks for Object Detection},'' in {\em IEEE Conference
  on Computer Vision and Pattern Recognition}, pp.~936--944, 2017.

\bibitem{Zhang_TII}
Y.~Zhang, M.~Liu, Y.~Yang, Y.~Guo, and H.~Zhang, ``{A Unified Light Framework
  for Real-Time Fault Detection of Freight Train Images},'' {\em IEEE
  Transactions on Industrial Informatics}, vol.~17, no.~11, pp.~7423--7432,
  2021.

\bibitem{DeepLab}
L.-C. {Chen}, G.~{Papandreou}, F.~{Schroff}, and H.~{Adam}, ``{Rethinking
  Atrous Convolution for Semantic Image Segmentation},'' {\em arXiv e-prints
  arXiv:1706.05587}, 2017.

\bibitem{125}
P.~Wang, P.~Chen, Y.~Yuan, D.~Liu, Z.~Huang, X.~Hou, and G.~Cottrell,
  ``{Understanding Convolution for Semantic Segmentation},'' in {\em IEEE
  Winter Conference on Applications of Computer Vision}, pp.~1451--1460, 2018.

\bibitem{GiOU}
H.~Rezatofighi, N.~Tsoi, J.~Gwak, A.~Sadeghian, I.~Reid, and S.~Savarese,
  ``{Generalized Intersection Over Union: A Metric and a Loss for Bounding Box
  Regression},'' in {\em IEEE/CVF Conference on Computer Vision and Pattern
  Recognition}, pp.~658--666, 2019.

\bibitem{IoU}
J.~Yu, Y.~Jiang, Z.~Wang, Z.~Cao, and T.~Huang, ``{UnitBox: An Advanced Object
  Detection Network},'' in {\em The 24th ACM International Conference on
  Multimedia}, pp.~516--520, 2016.

\bibitem{VGG}
K.~{Simonyan} and A.~{Zisserman}, ``{Very Deep Convolutional Networks for
  Large-Scale Image Recognition},'' {\em arXiv e-prints arXiv:1409.1556}, 2014.

\bibitem{InceptionV3}
C.~Szegedy, V.~Vanhoucke, S.~Ioffe, J.~Shlens, and Z.~Wojna, ``{Rethinking the
  Inception Architecture for Computer Vision},'' in {\em 2016 IEEE Conference
  on Computer Vision and Pattern Recognition}, pp.~2818--2826, 2016.

\bibitem{AdamW}
I.~{Loshchilov} and F.~{Hutter}, ``{Decoupled Weight Decay Regularization},''
  {\em arXiv e-prints arXiv:1711.05101}, 2017.

\bibitem{Zhaiyao_Zhang}
Y.~Zhang, K.~Lin, H.~Zhang, Y.~Guo, and G.~Sun, ``{A Unified Framework for
  Fault Detection of Freight Train Images Under Complex Environment},'' in {\em
  IEEE International Conference on Image Processing}, pp.~1348--1352, 2018.

\bibitem{caffe}
Y.~Jia, E.~Shelhamer, J.~Donahue, S.~Karayev, J.~Long, R.~Girshick,
  S.~Guadarrama, and T.~Darrell, ``{Caffe: Convolutional Architecture for Fast
  Feature Embedding},'' in {\em Proceedings of the 22nd ACM International
  Conference on Multimedia}, pp.~675--678, 2014.

\end{thebibliography}

\vspace{-10ex}
\begin{IEEEbiography}[
{\includegraphics[width=1in,height=1.25in,clip,keepaspectratio]{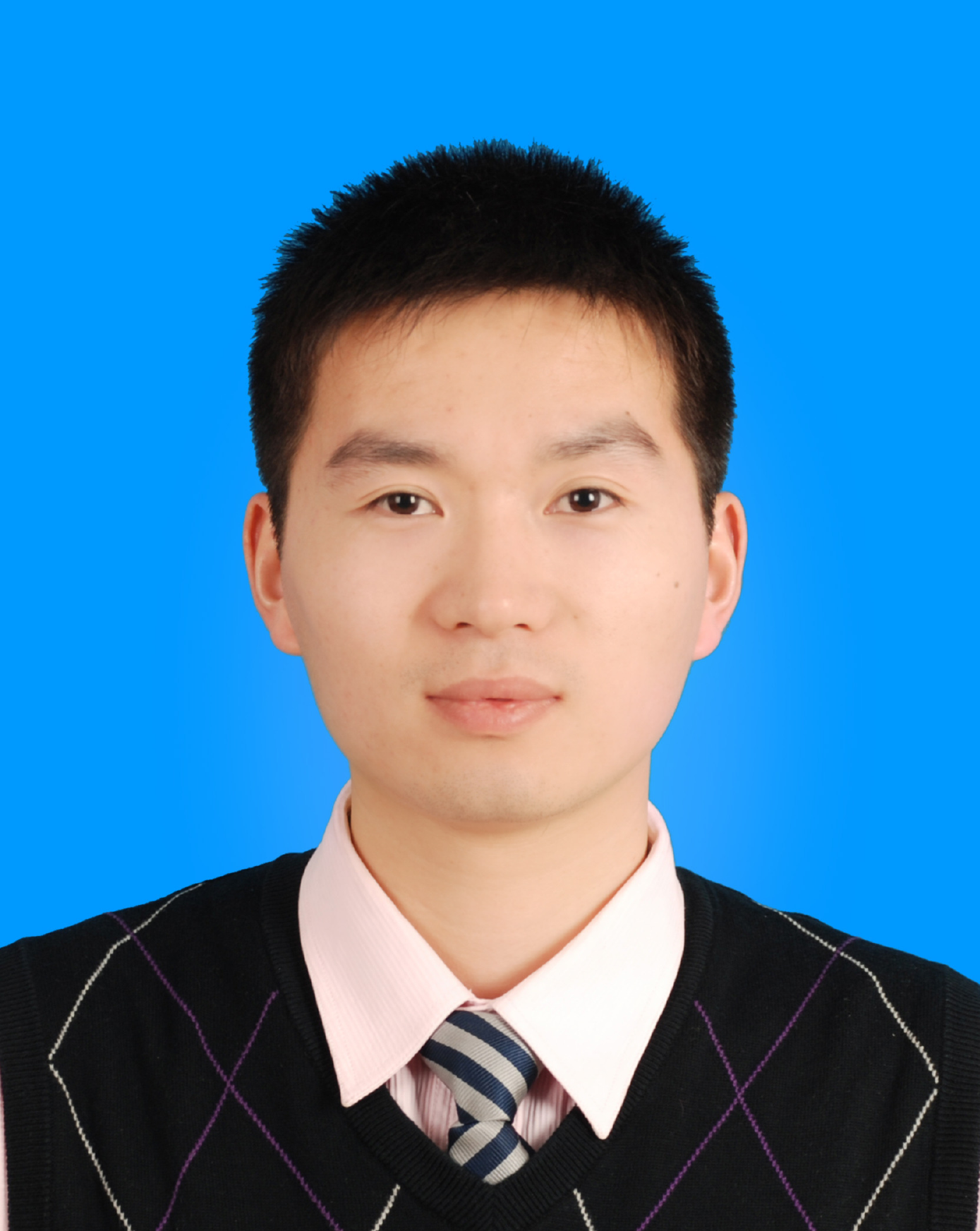}}]
{Guodong Sun}
is a professor in the School of Mechanical Engineering at the Hubei University of Technology. He received his BS degree in energy and power engineering and his PhD degree in mechanical and electronic engineering from Huazhong University of Science and Technology in 2002 and 2008, respectively. 

His current research interests include machine vision and imaging processing.
\end{IEEEbiography}

\vspace{-10ex}
\begin{IEEEbiography}[
{\includegraphics[width=1in,height=1.25in,clip,keepaspectratio]{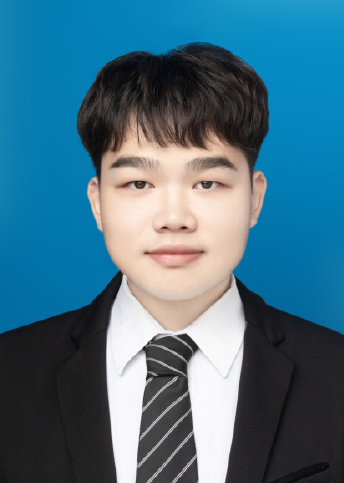}}]
{Yang Zhou}
received the B.S. degree in mechanical engineering from the Hubei University of Technology, Wuhan, China, in 2020. He is currently working toward the master's degree with the School of Mechanical Engineering, Hubei University of Technology, Wuhan.

His current research interests include computer vision and image processing.
\end{IEEEbiography}
\vspace{-30ex}
\begin{IEEEbiography}[
{\includegraphics[width=1in,height=1.25in,clip,keepaspectratio]{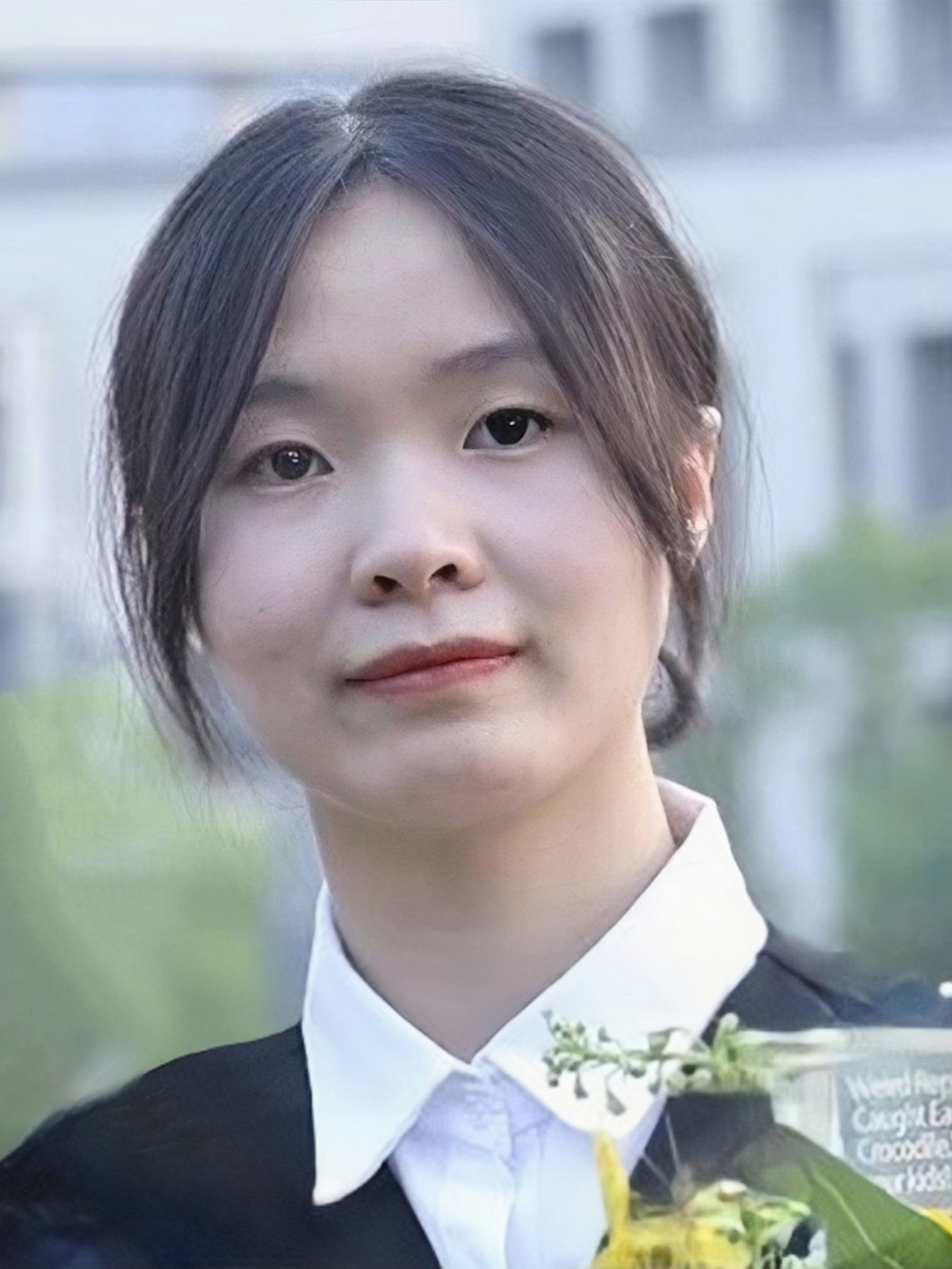}}]
{Huilin Pan}
received the B.S. degree in computer science and technology from the Anhui University of Technology, Maanshan, China, in 2021. She is currently working toward the M.S. degree with the mechanical engineering from the Hubei University of Technology, Wuhan, China. 

Her current research interests include machine learning and computer vision.

\end{IEEEbiography}
\vspace{-30ex}
\begin{IEEEbiography}[
{\includegraphics[width=1in,height=1.25in,clip,keepaspectratio]{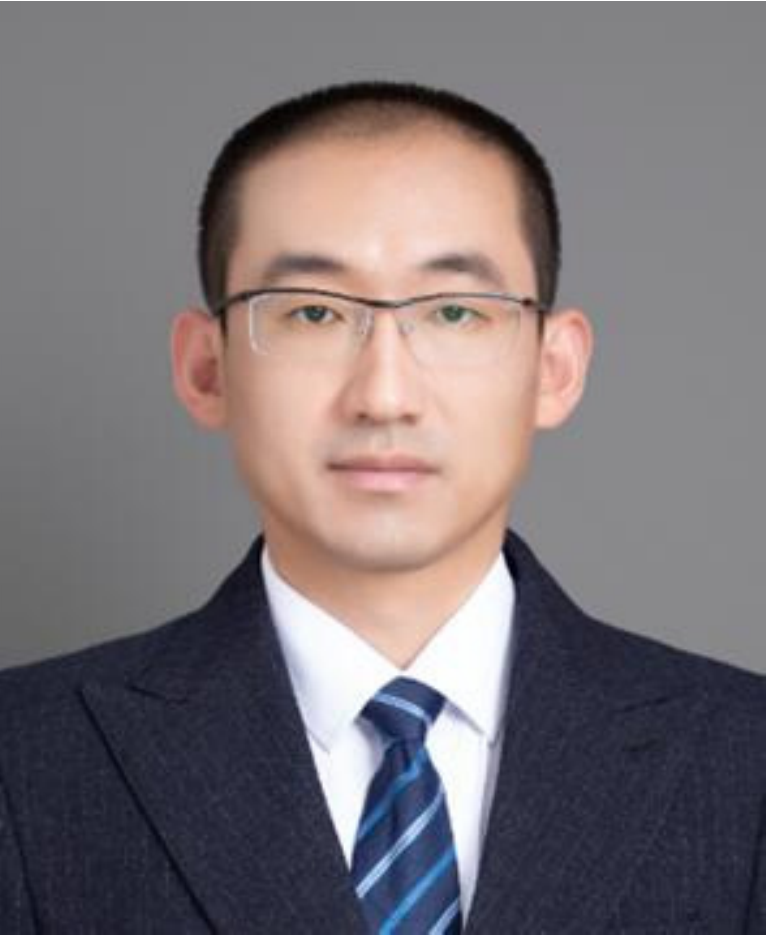}}]
{Bo Wu}
received Ph.D. degree from University of Chinese Academy of Sciences, Beijing, China, in 2019. Now he is a professor with Shanghai Advanced Research Institute. 

His current research interests are IoT, machine intelligence and fault diagnosis.
\end{IEEEbiography}

\vspace{-30ex}
\begin{IEEEbiography}[
{\includegraphics[width=1in,height=1.25in,clip,keepaspectratio]{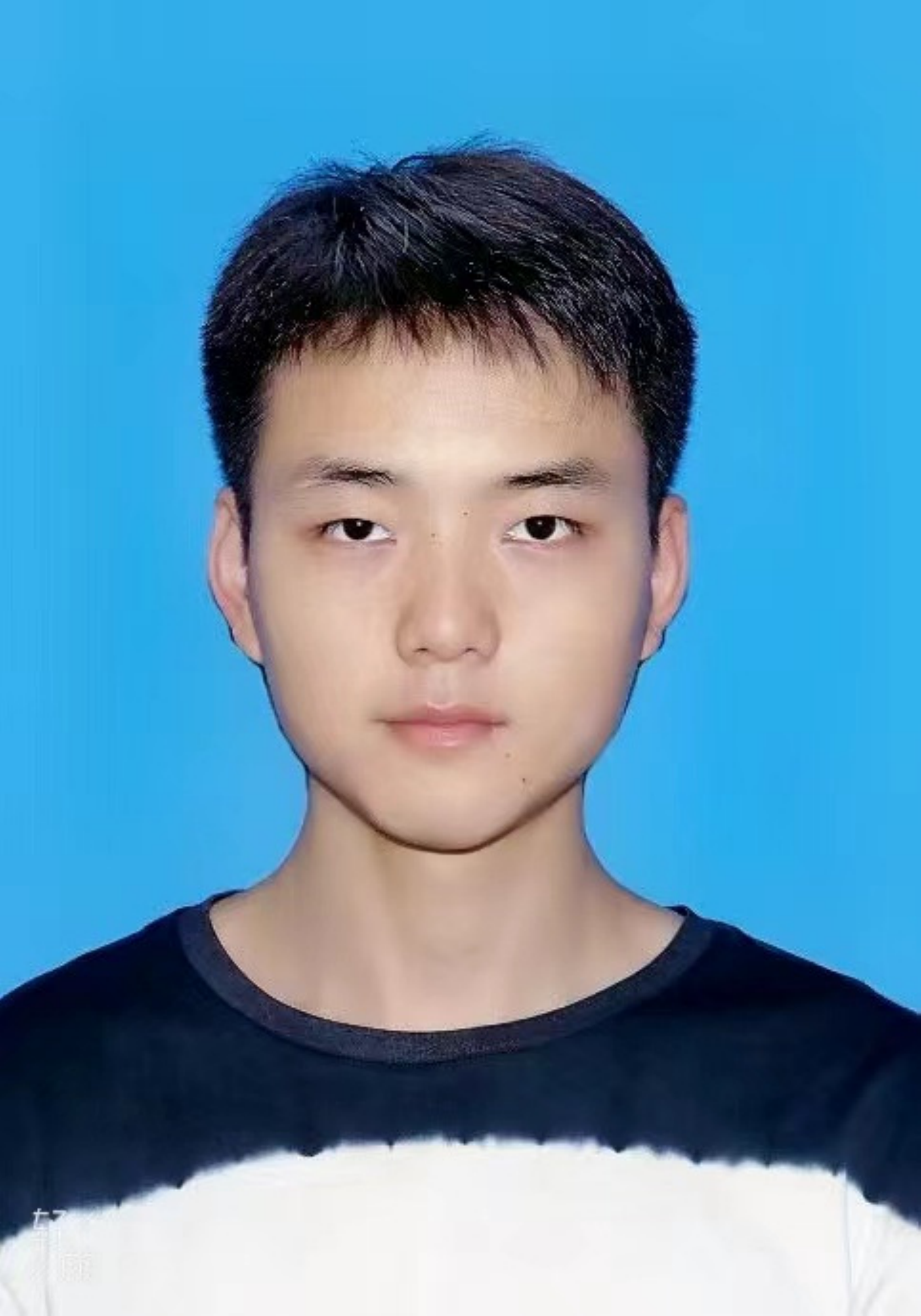}}]
{Ye Hu}
received the B.S. degree in mechanical engineering from the Hubei University of Technology, Wuhan, China, in 2020. He is currently working toward the master's degree with the School of Mechanical Engineering, Hubei University of Technology, Wuhan.

His current research interests include machine learning and signal processing.
\end{IEEEbiography}

\vspace{-30ex}
\begin{IEEEbiography}[
{\includegraphics[width=1in,height=1.25in,clip,keepaspectratio]{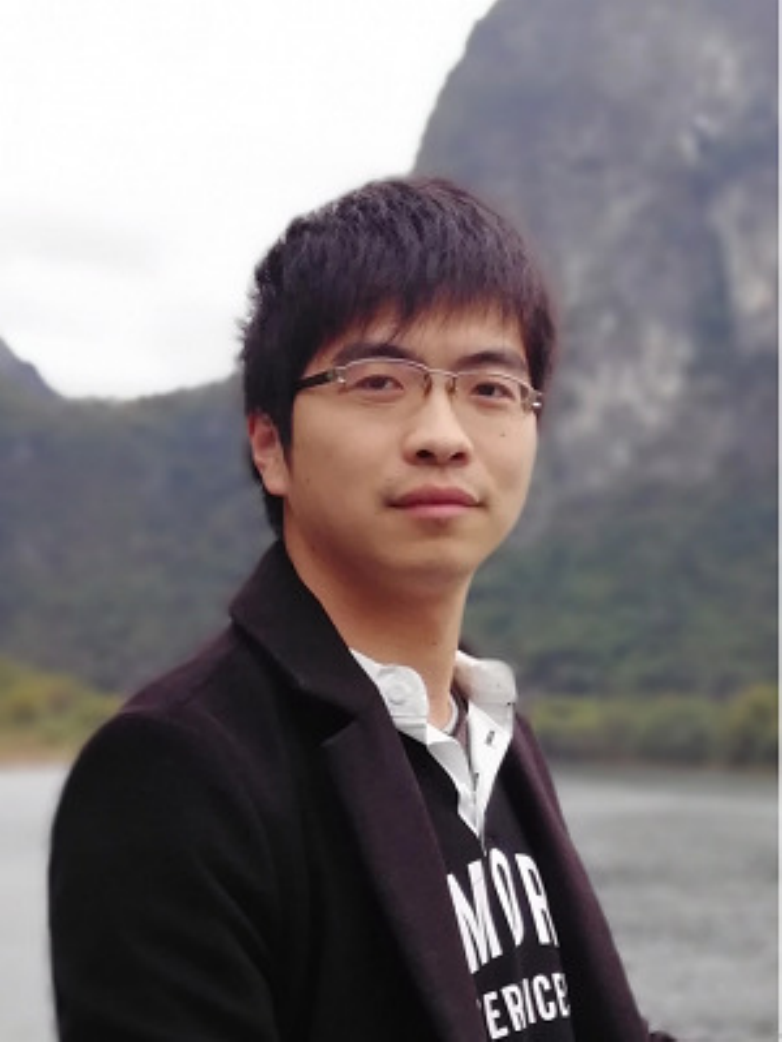}}]
{Yang Zhang}
received the Ph.D. degree in computer science and technology from the National Key Laboratory for Novel Software Technology, Department of Computer Science and Technology, Nanjing University, Nanjing, China, in 2021. He is currently an Assistant Professor with the School of Mechanical Engineering, Hubei University of Technology, Wuhan, China. 

His current research interests include machine learning and computer vision.
\end{IEEEbiography}

\end{document}